\documentclass[nohyperref]{article}

\usepackage{microtype}
\usepackage{graphicx}
\usepackage{subfigure}
\usepackage{booktabs} 

\usepackage{threeparttable}
\usepackage{makecell}
\usepackage{multirow}

\usepackage{colortbl}
\definecolor{bgcolor}{rgb}{0.94,0.97,1}
\definecolor{bgcolor2}{rgb}{0.8,1,0.8}
\definecolor{bgcolor3}{rgb}{0.50,0.90,0.50}

\usepackage[pagebackref=true,backref=true]{hyperref}
\renewcommand*{\backrefalt}[4]{%
    \ifcase #1 \footnotesize{(Not cited.)}%
    \or        \footnotesize{(Cited on page~#2)}%
    \else      \footnotesize{(Cited on pages~#2)}%
    \fi}

 \usepackage[accepted]{icml2022}

\usepackage{amsmath}
\usepackage{amssymb}
\usepackage{mathtools}
\usepackage{amsthm}
\usepackage{makecell}

\usepackage{nicefrac}

\usepackage{tcolorbox}
\usepackage{pifont}
\definecolor{mydarkgreen}{RGB}{39,130,67}
\definecolor{mydarkred}{RGB}{192,25,25}
\newcommand{\green}{\color{mydarkgreen}}
\newcommand{\red}{\color{mydarkred}}
\newcommand{\cmark}{\green\ding{51}}%
\newcommand{\xmark}{\red\ding{55}}%

\usepackage{xspace}
\usepackage{bm}

\usepackage{relsize} 
\newcommand{\algname}[1]{{\sf\green\relscale{0.90}#1}\xspace}

\usepackage[capitalize,noabbrev]{cleveref}

\theoremstyle{plain}
\newtheorem{theorem}{Theorem}[section]
\newtheorem{proposition}[theorem]{Proposition}
\newtheorem{lemma}[theorem]{Lemma}
\newtheorem{corollary}[theorem]{Corollary}
\theoremstyle{definition}
\newtheorem{definition}[theorem]{Definition}
\newtheorem{assumption}[theorem]{Assumption}
\theoremstyle{remark}
\newtheorem{remark}[theorem]{Remark}

\usepackage[colorinlistoftodos,bordercolor=orange,backgroundcolor=orange!20,linecolor=orange,textsize=scriptsize]{todonotes}

\newcommand{\seb}[1]{\todo[inline]{{\textbf{Sebastian:} \emph{#1}}}}
\newcommand{\peter}[1]{\todo[inline]{{\textbf{Peter:} \emph{#1}}}}

 \newcommand{\squeeze}{} 

\newcommand{\E}[1]{\mathbb{E}\left[#1\right]}
\newcommand{\Et}[1]{\mathbb{E}_t\left[#1\right]}

\newcommand{\cL}{\mathcal{L}}
\newcommand{\cO}{\mathcal O}
\newcommand{\cC}{\mathcal C}

\newcommand{\mI}{\mathbf{I}}

\newcommand{\mL}{\mathbf{L}}
\newcommand{\mW}{\mathbf{W}}
\newcommand{\R}{\mathbb R}

\renewcommand{\empty}{\varnothing}

\usepackage{mathtools}
\newcommand{\eqdef}{\coloneqq}

\newcommand{\avein}{\frac{1}{n}\sum_{i=1}^n}
\newcommand{\avemm}{\frac{1}{M}\sum_{m=1}^M}

\newcommand{\summm}{\sum_{m=1}^M}
\renewcommand{\phi}{\varphi}

\DeclareMathOperator{\prox}{prox}
\DeclareMathOperator{\argmin}{argmin}

\DeclareMathOperator{\range}{range}
\def\<#1,#2>{\langle #1,#2\rangle}

\icmltitlerunning{ProxSkip}

\begin{document}

\twocolumn[
\icmltitle{ProxSkip: Yes! Local Gradient Steps Provably Lead \\ to Communication Acceleration! Finally!${}^\dagger$}




\icmlsetsymbol{equal}{}

\begin{icmlauthorlist}
\icmlauthor{Konstantin Mishchenko}{inria}
\icmlauthor{Grigory Malinovsky}{kaust}
\icmlauthor{Sebastian Stich}{cispa}
\icmlauthor{Peter Richt\'{a}rik}{kaust}
\end{icmlauthorlist}

\icmlaffiliation{kaust}{Computer Science, King Abdullah University of Science and Technology, Thuwal, Saudi Arabia}
\icmlaffiliation{inria}{CNRS, ENS, Inria Sierra, Paris, France}
\icmlaffiliation{cispa}{CISPA Helmholtz Center for Information Security,  Saarbr\"{u}cken, Germany}

\icmlcorrespondingauthor{Peter Richt\'{a}rik}{peter.richtarik@kaust.edu.sa}

\icmlkeywords{Machine Learning, ICML}

\vskip 0.3in
]



\printAffiliationsAndNotice{}  

\begin{abstract}
We introduce \algname{ProxSkip}---a surprisingly simple and provably efficient method for minimizing the sum of a smooth ($f$) and an expensive nonsmooth proximable ($\psi$) function. The canonical approach to solving such problems is via the proximal gradient descent (\algname{ProxGD}) algorithm, which is based on the evaluation of the gradient of $f$ and the prox operator of $\psi$ in each iteration. In this work we are specifically interested in the regime in which the evaluation of prox is costly relative to the evaluation of the gradient, which is the case in many applications. \algname{ProxSkip} allows for the expensive prox operator to be skipped in most iterations: while its iteration complexity is $\mathcal{O}(\kappa \log \nicefrac{1}{\varepsilon})$, where $\kappa$ is the condition number of $f$, the number of prox evaluations is $\mathcal{O}(\sqrt{\kappa} \log \nicefrac{1}{\varepsilon})$ only. Our main motivation comes from federated learning, where evaluation of the gradient operator corresponds to taking a local \algname{GD} step independently on all devices, and evaluation of prox corresponds to (expensive) communication in the form of gradient averaging. In this context, \algname{ProxSkip} offers an effective {\em acceleration} of communication complexity. Unlike other local gradient-type methods, such as \algname{FedAvg}, \algname{SCAFFOLD}, \algname{S-Local-GD} and \algname{FedLin}, whose theoretical communication complexity is worse than, or at best matching, that of vanilla \algname{GD} in the heterogeneous data regime, we obtain a provable and large improvement without any heterogeneity-bounding assumptions. 
\end{abstract}

\section{Introduction}
\label{sec:intro}

We  study optimization problems of the  form
\begin{equation}
\min_{x\in \R^d} f(x) + \psi(x), \label{eq:main}
\end{equation}
where $f \colon \R^d\to \R$ is a smooth  function, and $\psi \colon \R^d\to \R \cup \{+\infty\}$ is a proper, closed and convex regularizer. 

Such problem are ubiquitous, and appear in numerous applications associated with virtually all areas of science and engineering, including signal processing \citep{ProxSplit2009}, image processing \citep{Luke2020}, data science \citep{Parikh2014:proxbook} and machine learning~\citep{shai_book}. 


\subsection{Proximal gradient descent}

One of the most canonical methods for solving \eqref{eq:main}, often used as the basis for further extensions and improvements, is proximal gradient descent (\algname{ProxGD}), also known as the forward-backward algorithm~\citep{ProxSplit2009,Nesterov_composite2013}.  This method solves \eqref{eq:main} via the iterative process defined by 
\begin{equation}\label{eq:ProxGD} x_{t+1} = \prox_{\gamma_t \psi} (x_t - \gamma_t \nabla f(x_t)),
\end{equation}
where $\gamma_t>0$ is a suitably chosen stepsize at time $t$, and $\prox_{\gamma \psi}(\cdot)\colon\R^d\to \R^d$ is the proximity operator of $\psi$, defined via
\begin{equation}\label{eq:prox}
\squeeze \prox_{\gamma \psi}(x) \eqdef \arg \min \limits_{y \in \R^d} \left[\frac{1}{2}\|y-x\|^2+ \gamma \psi(y)\right]. \end{equation}


It is typically assumed that the proximity operator \eqref{eq:prox} can be evaluated in closed form, which means that the iteration \eqref{eq:ProxGD} defining \algname{ProxGD} can be performed exactly.  \algname{ProxGD} is most suited to situations when the proximity operator is relatively cheap to evaluate, so that the bottleneck of  \eqref{eq:ProxGD} is in the forward step (i.e., computation of the gradient $\nabla f$) rather than in the backward step (i.e., computation of $\prox_{\gamma \psi}$). This is the case for many regularizers, including the $L_1$ norm ($\psi(x)=\|x\|_1$), the $L_2$ norm ($\psi(x) = \|x\|^2_2$), and elastic net \citep{ZouHastie:elastic-net:2005}. For many further examples, we refer the reader to the books~\citep{Parikh2014:proxbook,beck-book-first-order}.

\subsection{Expensive proximity operators}
However, in this work we are interested in the situation when the  evaluation of the {\em proximity operator is expensive}. That is, we assume that the computation of $\prox_{\gamma \psi}$ (the backward step)  is costly relative to the evaluation of the gradient of $f$ (the forward step). 

A conceptually simple yet rich class of expensive proximity operators arises from regularizers $\psi$ encoding a ``complicated-enough'' nonempty constraint set $\cC\subset \R^d$ via \begin{equation}\label{eq:indicator}\psi(x)=\begin{cases}0 & x \in \cC \\ +\infty & x\notin \cC\end{cases}.\end{equation}
The evaluation of the proximity operator of $\psi$ given by \eqref{eq:indicator} reduces to Euclidean projection onto $\cC$,
\[ \prox_{\gamma \psi}(x) = \arg \min_{y\in \cC} \|y-x\|, \]
 which can be a difficult optimization problem on its own. For instance, this is the case when $\cC$ is a polyhedral or a spectral set~\cite{Parikh2014:proxbook}.\footnote{Other examples of expensive proximity operators include Schatten-$p$ norms of matrices (e.g., the nuclear norm), and certain variants of quadratic support functions~\citep{QSF2016}.}




\subsection{Distributed machine learning and consensus constraints}

An important example of  expensive proximity operators associated with indicator functions \eqref{eq:indicator} arise in the {\em consensus} formulation of distributed optimization problems. In particular, consider the problem of minimizing the average of  $n$ functions using a cluster of  $n$ compute nodes/clients,
\begin{equation}\label{eq:finite-sum_097} \min_{x\in \R^d} \left\{f(x)\eqdef \frac{1}{n} \sum_{i=1}^n f_i(x) \right\}, \end{equation}
where function $f_i\colon\R^d \to \R$, and the data describing it, is owned by and stored on client $i \in [n]\eqdef \{1, 2, \dots, n\}$. This problem is of key importance in machine learning as it is an abstraction of the {\em empirical risk minimization} \citep{shai_book}, which is currently the dominant paradigm for training supervised machine learning models.

 By cloning the model $x\in \R^d$ into $n$ independent copies $x_1,\dots, x_n \in \R^d$, problem \eqref{eq:finite-sum_097} can be reformulated into the  {\em consensus form}~\citep[see e.g.][]{Parikh2014:proxbook}
\begin{equation}\label{eq:mainFL}
 \min_{x_1,\dots, x_n \in \R^d}  \frac{1}{n} \sum_{i=1}^n f_i(x_i) + \psi(x_1,\dots, x_n)\,,
\end{equation}
where the regularizer $\psi\colon\R^{nd}\to \R$ given by
\begin{equation}\label{eq:constraint_consensus_0980}
\psi(x_1,\dots, x_n) \eqdef 
	\begin{cases}
		0, & \textrm{if } x_1=\dotsb=x_n\,, \\
		+\infty, & \textrm{otherwise},
	\end{cases}
\end{equation}
encodes the consensus constraint \[\cC\eqdef \{(x_1,\dots,x_n) \in \R^{nd} \;:\; x_1=\dotsb=x_n\}.\]

Evaluating the proximity operator of \eqref{eq:constraint_consensus_0980} is not computationally expensive as it simply amounts to taking the average of the variables \citep{Parikh2014:proxbook}:
\begin{equation}\label{eq:prox-avg} \prox_{\gamma \psi}(x_1,\dots,x_n) = (\bar{x},\dots,\bar{x})\in \R^{nd}, \end{equation}
where
\begin{equation}\label{eq:avg}\bar{x} \eqdef \frac{1}{n}\sum_{i=1}^n x_i.\end{equation}
 However, it often involves \emph{high communication cost} since the vectors $x_1, \dots, x_n$  are stored on different compute nodes. Indeed, even simple averaging can be very time consuming if the communication links connecting the clients (e.g., through an orchestrating server) are slow and the dimension $d$ of the aggregated vectors/models high, which is the case in {\em federated learning (FL)}~\citep{FedLearn2016, Kairouz2019:federated}. 


\begin{table*}[!t]
	\centering
	\caption{The performance of federated learning methods employing multiple local gradient steps in the strongly convex regime.}\label{tbl:main-2}
	\begin{threeparttable}
		\footnotesize\setlength\tabcolsep{5.pt} 
		\begin{tabular}{ c  c  c  c  c  c  c }
			\toprule[.1em]
			 \begin{tabular}{c}\bf method \end{tabular} &  \begin{tabular}{c}\bf \# local steps \\  \bf per round \end{tabular} &\begin{tabular}{c}\bf \# floats sent  \\  \bf per round \end{tabular}   & \begin{tabular}{c}\bf stepsize \\ \bf on client $i$  \end{tabular}  & 
			\begin{tabular}{c} \bf linear \\ \bf rate? \end{tabular} & \begin{tabular}{c} \bf \# rounds \tnote{\color{blue}(c) }  \end{tabular}   & \begin{tabular}{c}\bf rate better  \\ \bf than \algname{GD}? \end{tabular} \\
			\midrule
			\begin{tabular}{c}\algname{GD}  {\tiny\citep{NesterovBook}}\end{tabular}  & $1$ & $d$&$\frac{1}{L}$ & \cmark & $\tilde{\cO}(\kappa)$  & \xmark \\ \cmidrule{1-1}
			\makecell{\algname{LocalGD} {\tiny\citep{khaled2019first, localSGD-AISTATS2020}}} & $\tau$  &$d$ & $\frac{1}{\tau L}$ & \xmark  & $\cO\left(\frac{G^2}{\mu n \tau \varepsilon}\right) $\tnote{\color{blue}(d)} & \xmark \\ \cmidrule{1-1}
			\makecell{\algname{Scaffold} {\tiny\citep{karimireddy2020scaffold}}} & $\tau$ & $2d$~\tnote{\color{blue}(e)}& $\frac{1}{\tau L}$ \tnote{\color{blue}(e)}& \cmark & $\tilde{\cO}(\kappa)$ & \xmark \\ \cmidrule{1-1}
			\makecell{\algname{S-Local-GD} \tnote{\color{blue}(a)}\quad {\tiny\citep{LSGDunified2020}}  }  & $\tau$ &$d<\#<2d$ \tnote{\color{blue}(f)}& $\frac{1}{\tau L}$  & \cmark & $\tilde{\cO}(\kappa)$  & \xmark \\ \cmidrule{1-1}
			\makecell{\algname{FedLin} \tnote{\color{blue}(b)}\quad {\tiny\citep{mitra2021linear}}} & $\tau_i$&$2d$  & $\frac{1}{\tau_i L} $ & \cmark &$\tilde{\cO}(\kappa)$  & \xmark \\
			\bottomrule[.1em] 
			\cellcolor{bgcolor} \begin{tabular}{c}\algname{Scaffnew}\tnote{\color{blue}(g)}\quad (this work) \\ for any $p\in (0,1]$ \end{tabular} & \cellcolor{bgcolor}$\frac{1}{p}$~\tnote{\color{blue}(h)} &\cellcolor{bgcolor}$d$&\cellcolor{bgcolor} $\frac{1}{L}$ & \cellcolor{bgcolor}\cmark & \cellcolor{bgcolor}$\tilde{\cO}\left(p \kappa + \frac{1}{p}\right)$ &\cellcolor{bgcolor} \begin{tabular}{c}{\cmark}\\ for $p \in \left(\frac{1}{\kappa},1\right)$\end{tabular} \\			
			\hline
			\cellcolor{bgcolor} \begin{tabular}{c}\algname{Scaffnew}\tnote{\color{blue}(g)}\quad  (this work) \\ for optimal $p=\frac{1}{\sqrt{\kappa}}$ \end{tabular} & \cellcolor{bgcolor}$\sqrt{\kappa}$~\tnote{\color{blue}(h)} &\cellcolor{bgcolor}$d$&\cellcolor{bgcolor} $\frac{1}{L}$ & \cellcolor{bgcolor}\cmark & \cellcolor{bgcolor}$\tilde{\cO}(\sqrt{\kappa})$ &\cellcolor{bgcolor} \cmark \\			
			\hline			
		\end{tabular} 
	\begin{tablenotes}
		{\scriptsize      
			\item [{\color{blue}(a)}] This is a special case of \algname{S-Local-SVRG}, which is a more general method presented in \citep{LSGDunified2020}. \algname{S-Local-GD} arises as a special case when full gradient is computed on each client.      
			\item [{\color{blue}(b)}]  \algname{FedLin} is a variant with a fixed but different number of local steps for each client. Earlier method \algname{S-Local-GD} has the same update but random loop length. 
			\item[{\color{blue}(c)}] The $\tilde{\cO}$ notation used in this column hides logarithmic factors.
			\item[{\color{blue}(d)}] $G$ is the level of dissimilarity at the solution $x_*$: $G^2 = \frac{1}{n}\sum_{i=1}^n \|\nabla f_i(x_*)\|^2$.
			\item[{\color{blue}(e)}] The number of floats used by Scaffold is $d$ when using Option II without client sampling. For the stepsize on client $i$, we use  \algname{Scaffold}'s cumulative local-global stepsize $\eta_l \eta_g$ for a  fair comparison.  
			\item[{\color{blue}(f)}] The number of sent vectors depends on hyper-parameters, and it is randomized.
			\item[{\color{blue}(g)}] \algname{Scaffnew} (Algorithm~\ref{alg:fl}) = \algname{ProxSkip} (Algorithm~\ref{alg:ProxSkip}) applied to the consensus formulation \eqref{eq:mainFL} + \eqref{eq:constraint_consensus_0980} of the finite-sum problem \eqref{eq:finite-sum_097}.
			\item[{\color{blue}(h)}] \algname{ProxSkip} (resp.\ \algname{Scaffnew}) takes a {\em random} number of gradient (resp.\ local) steps before  prox (resp.\ communication) is computed (resp.\ performed). What is shown in the table is the {\em expected} number of gradient (resp.\ local) steps.	}
	\end{tablenotes}  		
	\end{threeparttable}

\end{table*}

\subsection{Federated learning}

For the above reasons, practical FL algorithms generally use various communication-reduction mechanisms to achieve a useful computation-to-communication ratio, such as delayed communication. That is, the methods perform multiple local steps independently, based on their local objective~\cite{Mangasarian1994,mcdonald2010distributed,zhang2016parallel,%
McMahan16:FedLearning,stich2018local,lin2018don}.

However, when all the local functions $f_i$ are \emph{different} (i.e., when each individual machine has data drawn from a different distribution), local steps introduce
a {\em drift} in the updates of each client, which results in convergence issues. Indeed, even in the case of the simplest local gradient-type method, \algname{LocalGD}, a theoretical understanding that would not require any data similarity/homogeneity assumptions eluded the community for a long time. A resolution was found only recently~\citep{khaled2019first, localSGD-AISTATS2020, koloskova2020:unified}. However, the rates obtained in these works paint a pessimistic picture for \algname{LocalGD}; for example, due to client drift, they are sublinear even for smooth and  strongly convex problems. 

The next task for the FL community was to propose algorithmic adjustments that could provably mitigate the client drift issue. A   handful of recent methods, including
 \algname{Scaffold}~\cite{karimireddy2020scaffold}, \algname{S-Local-GD} \citep{LSGDunified2020} and \algname{FedLin}~\cite{mitra2021linear}, managed to do that.  For instance, under the assumption that $f$ is $L$-smooth and $\mu$-strongly convex, with condition number $\kappa = \nicefrac{L}{\mu}$, \algname{Scaffold}, \algname{S-Local-GD} and \algname{FedLin}  obtain a $\cO(\kappa \log \nicefrac{1}{\varepsilon})$ communication complexity, which matches the communication complexity of \algname{GD} (that computes a single gradient on every client per round of communication).  However, and despite the empirical superiority of these methods over vanilla \algname{GD}, their theoretical communication complexity does {\em not} improve upon \algname{GD}. This reveals a fundamental gap in our understanding of local methods.
 
Due to the enormous effort that was exerted over the last several years  by the FL community in this direction without it bearing the desired fruit~\cite{Kairouz2019:federated},  it seems very challenging to establish theoretically that performing independent local updates improves upon the communication complexity of \algname{GD}.  In contrast, accelerated gradient descent (without local steps)
 can reach the optimal  $\cO(\sqrt{\kappa} \log \nicefrac{1}{\varepsilon})$ communication complexity~~\cite{Lan2012:ac-sc,WoodworthPS20,woodworth21a}.
 
 This raises the question of whether this is a fundamental limitation of local methods. Is it possible to prove a better communication complexity than $\cO(\kappa \log \nicefrac{1}{\varepsilon})$ for {\em simple} local gradient-type methods, without resorting to any explicit acceleration mechanisms? 

\begin{algorithm*}[t]
	\caption{\algname{ProxSkip}}
	\label{alg:ProxSkip}
	\begin{algorithmic}[1]
		\STATE stepsize $\gamma > 0$, probability $p>0$, initial iterate $x_0\in \R^d$, initial control variate ${\red h_0} \in \R^d$, number of iterations $T\geq 1$
		\FOR{$t=0,1,\dotsc,T-1$}
		\STATE $\hat x_{t+1} = x_t - \gamma (\nabla f (x_t) - {\red h_t})$ \hfill $\diamond$ Take a gradient-type step adjusted via the control variate ${\red h_t}$
		\STATE Flip a coin $\theta_t \in \{0,1\}$ where $\mathop{\rm Prob}(\theta_t =1) = p$ \hfill $\diamond$ Flip a coin that decides whether to skip the prox or not
		\IF{$\theta_t=1$} 
		\STATE  $x_{t+1} = \prox_{\frac{\gamma}{p}\psi}\bigl(\hat x_{t+1} - \frac{\gamma}{p}{\red h_t} \bigr)$ \hfill $\diamond$ Apply prox, but only very rarely! (with small probability $p$)
		\ELSE
		\STATE $x_{t+1} = \hat x_{t+1}$ \hfill $\diamond$ Skip the prox!
		\ENDIF
		\STATE ${\red h_{t+1}} = {\red h_t} + \frac{p}{\gamma}(x_{t+1} - \hat x_{t+1})$ \hfill $\diamond$ Update the control variate ${\red h_t}$
		\ENDFOR
	\end{algorithmic}
\end{algorithm*}
\section{Contributions}
\label{sec:contributions}

We now summarize the main contributions of this work.

\subsection{\algname{ProxSkip}: a general prox skipping algorithm}
We develop a new \algname{ProxGD}-like algorithm for solving the general regularized problem \eqref{eq:main}. Our method,  which we call \algname{ProxSkip} (see Algorithm~\ref{alg:ProxSkip}), is designed to handle expensive proximal operators.  

A key ingredient in its design is a {\em randomized prox-skipping procedure}: in each iteration of \algname{ProxSkip}, we evaluate the proximity operator with probability $p \in (0,1]$. If $p=1$, several steps in our method are vacuous, and we recover \algname{ProxGD} as a special case (and the associated standard theory). 
Of course, the interesting choice  is $0<p<1$.  In expectation, the proximity operator is evaluated every $\nicefrac{1}{p}$ iterations, which can be very rare if $p$ is small.

{\bf Control variates stabilizing prox skipping.} We had  to introduce several new algorithmic design adjustments for such a  method to provably work.  In particular, \algname{ProxSkip} uses a control  variate $\red h_t$ on line 3 to shift the gradient $\nabla f(x_t)$ when the forward step is performed. 

Note that $\red h_t$ stays constant in between two consecutive prox calls. Indeed, this is because in that case we have $x_{t+1} = \hat x_{t+1}$  from line 8, and line 10 therefore simplifies to $\red h_{t+1} = \red h_t$.  So, when operating in between two prox calls, our method performs iterations of the form
\[x_{t+1} = x_t - \gamma (\nabla f (x_t) - {\red h_t}),\] where $\gamma>0$ is a stepsize parameter. 
When a prox step is executed, both the iterate $x_t$ and the control variate $\red h_t$ are adjusted, and the process is repeated.

This control mechanism  is necessary to allow for prox-skipping  to work.
To illustrate this, consider an optimal point $x_{\star} =\argmin_{x} f(x)+ \psi(x)$. In general, it does not hold $\nabla f(x_{\star})=0$, so skipping the prox (without control variate adjustment) would imply a  \emph{drift away} from $x_{\star}$. 
We show below that the control variate converges to 
\[{\red h_t} \to \nabla f(x_{\star}),
\]
which means that $x_{\star}$ is a  fixed point. This allows skipping the prox for a significant amount of steps without impacting the convergence.

{\bf Theory.} If $f$ is $L$-smooth and $\mu$-strongly convex, we prove that \algname{ProxSkip} converges at a linear rate. In particular, we show that after $T$ iterations, 
\[\E{\Psi_{T}}
		\le (1 - \min\{\gamma\mu, p^2\})^T \Psi_0,\]
where $\Psi_t$ is a certain Lyapunov function  (see \eqref{eq:Lyapunov_def}) involving both $x_{t}$ and $\red h_{t}$. If we choose $\gamma = \nicefrac{1}{L}$ and $p=\nicefrac{1}{\sqrt{\kappa}}$, where $\kappa = \nicefrac{L}{\mu}$ is the condition number, then the iteration complexity of \algname{ProxSkip}  is $\cO(\kappa\log\nicefrac{1}{\varepsilon})$, whereas the number of prox evaluations (in expectation) is $\cO(\sqrt{\kappa}\log\nicefrac{1}{\varepsilon})$ only! 
 For more details related to theory, see Section~\ref{sec:theory}.

\subsection{\algname{Scaffnew}: \algname{ProxSkip} applied to federated learning}

When applied to the consensus reformulation \eqref{eq:mainFL}--\eqref{eq:constraint_consensus_0980} of problem \eqref{eq:finite-sum_097}, \algname{ProxSkip} can be interpreted as a new distributed gradient-type method performing local steps, adding to the existing rich literature on local methods. 
In this context, we decided to call our method \algname{Scaffnew} (Algorithm~\ref{alg:fl}).\footnote{\scriptsize This is a homage to the influential \algname{Scaffold} method of \citet{karimireddy2020scaffold}, which in our experiments performs very similarly to \algname{Scaffnew} if the former method is used with fine-tuned stepsizes. } 
Since prox evaluation now means communication via averaging across the nodes (see \eqref{eq:prox-avg} and \eqref{eq:avg}), and since \algname{Scaffnew} inherits the strong theoretical prox-skipping properties of its parent method \algname{ProxSkip}:
\begin{quote}\em We resolve one of the most important open problems in the FL literature: breaking the $\cO(\kappa \log \nicefrac{1}{\varepsilon})$ communication complexity barrier with a simple local method. In particular,  \algname{Scaffnew} reaches an $\cO(\sqrt{\kappa} \log \nicefrac{1}{\varepsilon})$ communication complexity without imposing any additional assumptions (e.g., data similarity or stronger smoothness assumptions). 
\end{quote}

Note that since the iteration complexity of \algname{Scaffnew}  is $\cO(\kappa \log \nicefrac{1}{\varepsilon})$, the number of local steps per communication round is (on average)  $\cO(\sqrt{\kappa})$. According to~\citet{Arjevani2015:communication}, the communication lower bound for first order distributed algorithms is $\cO(\sqrt{\kappa} \log \nicefrac{1}{\varepsilon})$. This means that \algname{Scaffnew} is optimal in terms of communication rounds. \looseness=-1

Please refer to Table~\ref{tbl:main-2} in which we compare our results with the results obtained by existing state-of-the-art methods.

\subsection{Extensions}

We develop two extensions of the vanilla \algname{ProxSkip} method; see Section~\ref{sec:extensions}. We are not attempting to be exhaustive: these extensions are meant to illustrate that our method and proof technique combine well with other tricks and techniques often used in the literature. 

{\bf From deterministic to stochastic gradients.} First, in Section~\ref{sec:stochastic} we  perform an extension enabling us to use a {\em stochastic gradient} $g_{t}(x_t)\approx \nabla f(x_t)$ in  \algname{ProxSkip}  instead of the true gradient $\nabla f(x_t)$.  This is of importance in many applications, and is of particular importance for our method since now that the cost of the prox step was reduced, the cost of the gradient steps becomes more important. We operate under the modern {\em expected smoothness} assumption introduced by \citet{gower2019sgd,gower2021stochastic}, which is less restrictive than the standard bounded variance assumption.

{\bf From a central server to fully decentralized training.} Second, in Section~\ref{sec:decentralized} we present and analyze  \algname{ProxSkip} in a fully {\em decentralized} optimization setting, where the communication between nodes is restricted to a communication graph.
Our decentralized algorithm inherits the  property that it is not affected by data-heterogeneity. 
The covariate technique in \algname{ProxSkip} resembles, to some extent, some of the existing \emph{gradient tracking} mechanisms~\citep{Lorenzo2016GT-first-paper,Nedic2016DIGing}.  However, while gradient tracking provably addresses data-heterogeneity, its communication complexity scales proportional to the iteration complexity, $\cO(\kappa)$~\citep{Yuan2021d2-exact-diff-rates,koloskova2021improved}.
The same holds for almost all other schemes that have been designed to address data-heterogeneity in decentralized optimization \citep{Tang2018:d2,Vogels2021:relay}. Notable exceptions include the optimal methods developed by \citet{OPAPC,ADOM,ADOM+}; see also the references therein. However, these methods are based on classical acceleration schemes, and do not perform multiple local steps.

\section{Theory}
\label{sec:theory}

We are now ready to describe our key theoretical development: the convergence analysis of \algname{ProxSkip}.

\subsection{Assumptions}
We rely on several standard  assumptions to establish our results. First, we need $f$ to be smooth and strongly convex (see Appendix~\ref{appendix:facts} for complementary details).
\begin{assumption}\label{as:f}
	$f$ is $L$-smooth and $\mu$-strongly convex.
	\end{assumption}

We also need the following standard assumption\footnote{\scriptsize Note that this assumption is automatically satisfied for $\psi$ defined in~\eqref{eq:constraint_consensus_0980}.} on the regularizer $\psi$.

\begin{assumption}\label{as:proper_psi}
$\psi$ is proper, closed and convex.
\end{assumption}

These assumption imply that problem \eqref{eq:main} has a unique minimizer, which we denote $x_{\star}\eqdef \argmin f(x)+\psi(x)$.

\subsection{Firm nonexpansiveness}
In one step of our analysis we will rely on firm nonexpansiveness of the proximity operator \citep[see, e.g.,][]{bauschke2021generalized}: \begin{lemma}\label{lem:prox-contraction} 
	Let \Cref{as:proper_psi} be satisfied. Let $P(x)\eqdef \prox_{\frac{\gamma}{p}\psi}(x)$ and $Q(x) \eqdef x - P(x)$. Then 
	\begin{equation}
		\|P(x)-P(y)\|^2 + \|Q(x) - Q(y)\|^2
		\le \|x - y\|^2, \label{eq:prox_firm_non_exp}
	\end{equation}
	for all $x, y \in \R^d$ and any $\gamma,p>0$.
\end{lemma}

\subsection{Two technical lemmas}

The strength of our method comes from the role the control variates $\red h_t$ play in stabilizing the effect of skipping prox evaluations. Our analysis captures this effect. In particular, a by-product of our analysis is a proof that the control variates converge to  
$h_{\star}\eqdef \nabla f(x_{\star})$, where $x_{\star}$ is the solution. In order to show this, we work with the following natural candidate for a Lyapunov function: \begin{equation} \label{eq:Lyapunov_def} \Psi_t \eqdef \|x_{t} - x_{\star}\|^2 + \frac{\gamma^2}{p^2}\|h_{t} - h_{\star}\|^2\,.\end{equation}
We  further define
\begin{equation}\label{eq:98g9gbjfd8d}\squeeze w_t \eqdef x_t - \gamma \nabla f(x_t), \quad \text{and} \quad w_{\star} \eqdef x_{\star} - \gamma \nabla f(x_{\star}).\end{equation} 
Note that if our method works, i.e., if $x_t\to x_{\star}$, then gradient smoothness  implies that $w_t \to w_{\star}$. In our first technical lemma, we show that after one step of \algname{ProxSkip}, the Lyapunov function can be bounded in terms of the distance $\|w_t - w_{\star}\|^2$ and the control variate error $\|h_{t} - h_{\star}\|^2$. It is this lemma in the proof of which we rely on firm nonexpansiveness. We do not use it anywhere else.
	
\begin{lemma}\label{lem:A}
If Assumptions \ref{as:f} and \ref{as:proper_psi} hold, $\gamma >0$ and $0<p\leq 1$, then
\begin{equation} \label{eq:b8f9d89fd8df_09}\squeeze \E{ \Psi_{t+1} } \leq  \|w_t - w_{\star}\|^2 +  (1-p^2)\frac{\gamma^2}{p^2}\|h_{t} - h_{\star}\|^2 \,, \end{equation}
where the expectation is taken over the $\theta_t$ in Algorithm~\ref{alg:ProxSkip}.
\end{lemma}

 Our next lemma bounds the first term in the right-hand side of \eqref{eq:b8f9d89fd8df_09} by a multiple of $\|x_t-x_{\star}\|^2$.
\begin{lemma} \label{lem:B} Let \Cref{as:f} hold with any $\mu\geq 0$. If $0< \gamma \leq \frac{1}{L}$, then \begin{equation}\label{eq:nbo98fd8f_09uf}\|w_{t} - w_{\star}\|^2 \le (1-\gamma\mu)\|x_t - x_{\star}\|^2.\end{equation}
\end{lemma}

\subsection{Main theorem}

As we shall now see, our main theorem follows simply by combining the last two lemmas.

\begin{algorithm*}[t]
	\caption{\algname{Scaffnew}: Application of \algname{ProxSkip} to Federated Learning (i.e., to problem \eqref{eq:mainFL}--\eqref{eq:constraint_consensus_0980})}
	\label{alg:fl}
	
	\let\oldwhile\algorithmicwhile
	\renewcommand{\algorithmicwhile}{\textbf{in parallel on all workers $i \in [n]$}}
	\let\oldendwhile\algorithmicendwhile
	\renewcommand{\algorithmicendwhile}{\algorithmicend\ \textbf{local updates}}
	
	\begin{algorithmic}[1]
		\STATE stepsize $\gamma > 0$, probability $p>0$, initial iterate $x_{1,0}=\dots = x_{n,0}\in \R^d$, initial control variates ${\red h_{1,0}, \dots, h_{n,0}}  \in \R^d$  on each client such that  $\sum_{i=1}^{n}{\red h_{i,0}} = 0$, number of iterations $T\geq 1$
		\STATE \textbf{server:} flip a coin, $\theta_t \in \{0,1\}$, $T$ times,  where $\mathop{\rm Prob}(\theta_t =1) = p$\hfill $\diamond$ Decide when to skip communication
		\STATE send the sequence  $\theta_0, \dots, \theta_{T-1}$ to all workers 
		\FOR{$t=0,1,\dotsc,T-1$}
		\WHILE{}
		\STATE $\hat x_{i,t+1} = x_{i,t} - \gamma (g_{i,t} (x_{i,t}) - {\red h_{i,t}})$ \hfill $\diamond$ Local gradient-type step adjusted via the local control variate ${\red h_{i,t}}$
		\IF{$\theta_t=1$} 
		\STATE  $x_{i,t+1} = \frac{1}{n}\sum \limits_{i=1}^n \hat x_{i,t+1}$ \hfill  $\diamond$ Average the iterates, but only very rarely! (with small probability $p$)
		\ELSE
		\STATE $x_{i,t+1} = \hat x_{i,t+1}$ \hfill $\diamond$ Skip communication!
		\ENDIF
		\STATE ${\red h_{i,t+1}} = {\red h_{i,t}} + \frac{p}{\gamma}(x_{i,t+1} - \hat x_{i,t+1})$ \hfill $\diamond$ Update the local control variate ${\red h_{i,t}}$
		\ENDWHILE
		\ENDFOR
	\end{algorithmic}
\end{algorithm*}

\label{sec:maintheorem}
\begin{theorem}
 \label{thm:main}
	Let \Cref{as:f} and \Cref{as:proper_psi} hold, and let $0<\gamma \le \frac{1}{L}$ and $0<p\leq 1$. Then, the iterates of \algname{ProxSkip} (\Cref{alg:ProxSkip}) satisfy
	\begin{equation}\label{eq:lyapunov_decrease}
		\E{\Psi_{T}}
		\le (1 - \zeta)^T \Psi_0,
	\end{equation}
	where $\zeta\eqdef \min\{\gamma\mu, p^2\}$.
\end{theorem}

\begin{proof}
By combining Lemmas~\ref{lem:A} and \ref{lem:B}, we get
	\begin{align*}
		\E{\Psi_{t+1}}
				&\le (1 - \gamma\mu)\|x_t - x_{\star}\|^2 + (1-p^2)\frac{\gamma^2}{p^2}\|h_{t} - h_{\star}\|^2 \\
		&\le (1-\zeta)\left(\|x_t - x_{\star}\|^2 + \frac{\gamma^2}{p^2}\|h_{t} - h_{\star}\|^2 \right) \\
		& = (1-\zeta) \Psi_t.
	\end{align*}
	We get the theorem's claim by unrolling the recurrence.
\end{proof}

\subsection{How often should one skip the prox?}

Note that by choosing $p=1$ (no prox skipping) and $\gamma=\nicefrac{1}{L}$, we get $\zeta=\nicefrac{1}{\kappa}$, which leads to  the rate $\cO(\kappa \log \nicefrac{1}{\varepsilon})$ of \algname{ProxGD}. This is not a surprise since when $p=1$, \algname{ProxSkip} {\em is} identical to \algname{ProxGD}. 

More importantly, note that for any fixed stepsize $\gamma>0$, the reduction factor $\zeta\eqdef \min\{\gamma\mu, p^2\}$ in \eqref{eq:lyapunov_decrease} remains unchanged as we decrease $p$ from $1$ down to $p=\nicefrac{1}{\sqrt{\gamma \mu}}$. This is the reason why we can often {\em skip the prox}, and get away with it {\em for free}, i.e., without any deterioration of the convergence rate! 

 By inspecting \eqref{eq:lyapunov_decrease} it is easy to see that \begin{equation}\label{eq:nbi9fgd9gfd9_90}T\geq \max \left\{\frac{1}{\gamma \mu}, \frac{1}{p^2}\right\} \log \frac{1}{\varepsilon} \; \Longrightarrow \; \E{\Psi_T} \leq \varepsilon \Psi_0.\end{equation} Since in each iteration we evaluate the prox with probability $p$, the {\em expected number of 
prox evaluations} is \begin{equation}\label{eq:pT}pT \overset{\eqref{eq:nbi9fgd9gfd9_90}}{\approx}  \max \left\{\frac{p}{\gamma \mu},\frac{1}{p} \right\}\log \frac{1}{\varepsilon}.\end{equation}
Clearly, the best result is obtained if we use the largest stepsize allowed by \Cref{thm:main}: \begin{equation}\label{eq:niubfd_0909}\gamma =\frac{1}{L}.\end{equation} 
Next, the value of $p$ that minimizes expression \eqref{eq:pT} satisfies $\frac{pL}{ \mu}=\frac{1}{p}$, which gives the {\em optimal probability} \begin{equation}\label{eq:iuufd7-72332}p=\sqrt{\frac{\mu}{L}}=\frac{1}{\sqrt{\kappa}},\end{equation}
where $\kappa\eqdef \nicefrac{L}{\mu}$ is the condition number. With these optimal choices of the parameters $\gamma$ and $p$, the number of iterations of \algname{ProxSkip} is 
\[T\overset{\eqref{eq:nbi9fgd9gfd9_90}}{\approx}  \max\left\{\frac{1}{\gamma \mu} , \frac{1}{p^2}\right\}\log \frac{1}{\varepsilon} \overset{\eqref{eq:niubfd_0909}+\eqref{eq:iuufd7-72332}}{=}  \kappa \log \frac{1}{\varepsilon},
\]
and the expected number of prox evaluations performed in the process is 
\[
pT \overset{\eqref{eq:pT}}{\approx}   \max \left\{\frac{p}{\gamma \mu},\frac{1}{p} \right\}\log \frac{1}{\varepsilon} \overset{\eqref{eq:niubfd_0909}+\eqref{eq:iuufd7-72332}}{=} 
\sqrt{\kappa}\log \frac{1}{\varepsilon}.
\]

 Let us summarize the above findings.

\begin{corollary}
	If we choose $\gamma = \nicefrac{1}{L}$ and $p=\nicefrac{1}{\sqrt{\kappa}}$, then the iteration complexity of \algname{ProxSkip} (\Cref{alg:ProxSkip}) is $\cO(\kappa\log\nicefrac{1}{\varepsilon})$ and its prox calculation complexity is $\cO(\sqrt{\kappa}\log\nicefrac{1}{\varepsilon})$.
\end{corollary}


\section{Application to Federated Learning}
Let us now consider the problem of minimizing 
the average of $n$ functions stored on $n$ devices, as formulated in~\eqref{eq:finite-sum_097}.
This is the canonical problem in federated learning~\cite{McMahan16:FedLearning,Kairouz2019:federated}.\footnote{\scriptsize
As our focus is on a new communication-efficient scheme, we  disregard here other important aspects such as client sampling.} 
In this setting the functions $f_i \colon \R^d \to \R$ denote the local loss function of client $i$ defined over its own private data. For simplicity, we assume in this section that every client can compute the gradient $\nabla f_i(x)$ exactly (i.e., a full pass over the local data), see Section~\ref{sec:stochastic} for the discussion of the stochastic setting. When applied to the consensus reformulation \eqref{eq:mainFL}--\eqref{eq:constraint_consensus_0980} of problem \eqref{eq:finite-sum_097}, \algname{ProxSkip} reduces to \algname{Scaffnew} (Algorithm~\ref{alg:fl}).

{\bf Method description.} 
Algorithm~\ref{alg:fl} has three main steps:\ local updates  to the client model $x_{i,t} \in \R^d$, local updates to the client control variate ${\red h_{i,t}}\in\R^d$, and averaging the client models with probability $p$ in every iteration.

When $g_{i,t}(x_{i,t}) = \nabla f_i(x_{i,t})$, then each local update on client $i$ takes the form
\begin{align*}
 \squeeze\hat x_{i,t+1} = x_{i,t} - \gamma(\nabla f_i(x_{i,t}) - {\red h_{i,t}})\,.
\end{align*}
We will show below that ${\red h_{i,t} } \stackrel{t \to \infty}{\to} \nabla f_i(x_{\star}),$ so that it becomes evident that the optimal solution $x_\star$ is a fixed point of the algorithm (this is a key difference from, e.g., \algname{LocalGD} \citep{localSGD-AISTATS2020, koloskova2020:unified,malinovskiy2020local}). The local covariates ${\red h_{i,t}}$ are updated after each communication round, i.e., when $\theta_t=1$, as
\begin{align*}
 {\red h_{i,t+1}} = {\red h_{i,t}} + \frac{p}{\gamma} \underbrace{\left(\frac{1}{n}\sum_{j=1}^n \hat x_{j,t+1}  - \hat x_{i,t+1} \right)}_{\text{accumulated `client drift'}}\,.
\end{align*}
The local drift (i.e., deviation from the client mean) is divided by the stepsize and the expected length (i.e., $\nicefrac{1}{p}$) of the local phase during which the drift has been accumulated. 
This drift correction shares similarities with option II in \algname{Scaffold}~\cite{karimireddy2020scaffold} and \algname{QG-DSGD}~\cite{lin2021quasiglobal}, yet differs from option I in \algname{Scaffold} and \algname{FedLin}~\cite{mitra2021linear}, that both propose to compute an additional gradient at the client average.


\subsection{Convergence}

We will need an assumption on the individual functions $f_i$:

\begin{assumption}\label{as:f_i}
 Each $f_i$ is $L$-smooth and $\mu$-strongly convex.
\end{assumption}

Note though that we do not need to make any assumption on the similarity of the functions $f_i$. Convergence of  \algname{Scaffnew} (Algorithm~\ref{alg:fl}) in the deterministic case follows as a corollary of Theorem~\ref{thm:main}. 

\begin{corollary}[Federated Learning]
Let Assumption~\ref{as:f_i} hold and let $\gamma = \nicefrac{1}{L}$, $p=\nicefrac{1}{\sqrt{\kappa}}$ and $g_{i,t}(x_{i,t}) = \nabla f_i(x_{i,t})$. Then the iteration complexity of Algorithm~\ref{alg:fl} is $\cO(\kappa \log \nicefrac{1}{\varepsilon}$) and its communication complexity is $\cO(\sqrt{\kappa}\log \nicefrac{1}{\varepsilon})$.
\end{corollary}

{\bf Usefulness of local steps.}
Our result shows for the first time a real advantage of local update methods \emph{without imposing any similarity assumptions}. For instance, \citet{woodworth2020:local} assume quadratic functions, \citet{karimireddy2020scaffold,karimireddy2020:mime} bounded Hessian dissimilarity, and \citet{Yuan2020:accelerated} bounded Hessian. Without any such assumption, we show here that local methods can converge in significantly fewer update rounds than large-batch methods without local steps~\cite{Dekel2012:minibatch}. The method matches the  communication-complexity lower bound derived in~\cite{Arjevani2015:communication} and is optimal in this regard.  Moreover, and unlike the approach adopted by \citet{FL-personal-mixture2020,personalized-optimal-2020},  our improvements do not rely on interpreting local methods as methods for solving {\em personalized formulations of FL}.

%

\section{Extensions} \label{sec:extensions}

\subsection{Stochastic gradients}
\label{sec:stochastic}
In machine learning, calculating full gradients may be extremely expensive and in some cases not possible. In this section, we are going to make an extension of the basic \algname{ProxSkip}(\Cref{alg:ProxSkip}) to allow stochastic updates:
\begin{align}
\hat x_{t+1} = x_{t} - \gamma ({\color{blue} g_{t} (x_{t})} - h_{t}).
\end{align}
In a generic \algname{SGD} method, we  work with unbiased estimators of gradients only.
\begin{assumption}[Unbiasedness]\label{as:unbias}
For all $t \geq 0$, $g_t(x_t)$ is an unbiased estimator of the gradient $\nabla f(x_{t})$. That is,
\begin{align} 
\E{g_{t}(x_{t}) \mid x_{t}} =\nabla f(x_{t}).
\end{align}
\end{assumption}

In our analysis of \algname{ProxSkip} in the stochastic case, we rely on the {\em expected smoothness} assumption introduced by~\citet{gower2021stochastic} in the context of variance reduction, and later adopted and simplified by \citet{gower2019sgd} in the context if \algname{SGD} analysis.
\begin{assumption}[Expected smoothness]\label{as:exp_smooth}
	There exist constants $A \geq 0$ and $C \geq 0$ such that for all $t \geq 0$,
	\begin{align}
	 \mathrm{E}\left[\left\|g_{t}(x_{t})-\nabla f\left(x_{\star}\right)\right\|^{2} \mid x_{t}\right] \leq 2 A D_{f}\left(x_{t}, x_{\star}\right)+C.
\end{align}
\end{assumption}
This assumption is satisfied in many practical settings, including when the randomness in $g_{t}$ arises from subsampling (i.e., minibatching) and compression~\citep{LSGDunified2020}. It is also satisfied in the popular but artificial setting when an additive zero mean and bounded variance noise is added to the gradient, formalized next.
\begin{assumption}[Bounded variance]\label{as:bounded}
For all $t \geq 0$, the stochastic estimator $g_t(x_t)$ has bounded variance:
	\begin{align}
\operatorname{Var}[g_t(x_t) \;|\; x_t] \leq \sigma^{2}.
	\end{align}
\end{assumption}
The next lemma, due to \citet{gower2019sgd}, shows that this is indeed the case. \begin{lemma}
	\label{lemma:exp_smooth}
	Let~\Cref{as:unbias} and~\Cref{as:bounded} hold and let $f$ be convex and $L$-smooth, then expected smoothness (i.e.,~\Cref{as:exp_smooth}) holds with $A = L$ and $C = \sigma^2$.
\end{lemma}

\begin{algorithm*}[t]
	\caption{\algname{Decentralized Scaffnew}}
	\label{alg:dist_gd}
	\begin{algorithmic}[1]
		\STATE stepsizes $\gamma > 0$ and $\tau>0$, initial iterates $x_{1,0} = \ldots = x_{n,0} = x_0 \in \R^d$, initial control variables ${\red h_{1,0}} = \ldots = {\red h_{n,0}} = 0\in\R^d$, weights for averaging $\mathbf{W}=(W_{ij})_{i,j=1}^n$
		\FOR{$t=0,1,\dotsc,T-1$}
		\STATE Flip a coin $\theta_t \in \{0,1\}$ where $\mathop{\rm Prob}(\theta_t =1) = p$ \hfill $\diamond$ Flip a coin that decides whether to skip the prox or not
		\FOR{$i=1,\dotsc, n$}
		\STATE $\hat x_{i,t+1} = x_{i,t} - \gamma (\nabla f_i (x_{i,t}) - {\red h_{i,t}})$\hfill $\diamond$ Take a gradient-type step adjusted via the control variate ${\red h_{i,t}}$
		\IF{$\theta_t=1$}
		\STATE $x_{i,t+1} = \left(1 - \frac{\gamma\tau}{p}\right)\hat x_{i,t+1} + \frac{\gamma\tau}{p}\sum_{j=1}^n W_{ij}\hat x_{j,t+1}$ \hfill $\diamond$  Communicate, but only very rarely! (with small prob.\ $p$)
		\STATE ${\red h_{i,t+1}} = {\red h_{i,t}} + \frac{p}{\gamma}(x_{i,t+1} -  \hat x_{i, t+1})$  \hfill $\diamond$ Update the control variate ${\red h_{i,t}}$
		\ELSE
		\STATE $x_{i,t+1} = \hat x_{i, t+1}$ \hfill$\diamond$ Skip communication!
		\STATE ${\red h_{i,t+1}} =  {\red h_{i,t}}$
		\ENDIF
		\ENDFOR
		\ENDFOR
	\end{algorithmic}
\end{algorithm*}
The main result of this section is formulated next.
\begin{theorem}
	\label{thm:main-stoch}
	Let Assumptions~\ref{as:f}, \ref{as:proper_psi}, \ref{as:exp_smooth} and \ref{as:unbias} hold. Let $0<\gamma \le \nicefrac{1}{A}$ and $0<p\leq 1$. Then, the iterates of \algname{SProxSkip}(\Cref{alg:stoch_rand_prox}) satisfy
	\begin{align*}
		\squeeze\E{\Psi_{T}}
		\le (1 - \zeta)^T \Psi_0 + \frac{\gamma^2 C}{\zeta},
	\end{align*}
	where $\zeta\eqdef \min\{\gamma\mu, p^2\}$.
\end{theorem}
This result also gives us rates for~\algname{Scaffnew}(\Cref{alg:fl}).
\begin{corollary}\label{cor:0099887766}
 Consider  \algname{Scaffnew} (\Cref{alg:fl}) or \algname{SProxSkip}(\Cref{alg:stoch_rand_prox}). Choose any $0<\varepsilon <1$. If we choose $\gamma = \min\left\{  \frac{1}{A}, \frac{ \varepsilon \mu}{2 C} \right\}$ and $p=\sqrt{\gamma \mu}$,  then in order to guarantee $\E{ \Psi_{T} } \leq \varepsilon   $, it suffices to take
	$T \geq \max \left\{ \frac{A}{\mu}, \frac{2C}{\varepsilon \mu^2}\right\}\log \left(\frac{2 \Psi_{0}}{\varepsilon}\right)$ iterations, which results in $\max \left\{ \sqrt{\frac{A}{\mu}}, \sqrt{\frac{2C}{\varepsilon \mu^2}}\right\}\log \left(\frac{2 \Psi_{0}}{\varepsilon}\right)$ communications (in case of \algname{Scaffnew}) resp.\ prox evaluations (in case of  \algname{SProxSkip}) on average.
\end{corollary}
\begin{figure*}[t]
	\centering
	\begin{tabular}{ccc}
		$\!\!$\includegraphics[scale=0.25]{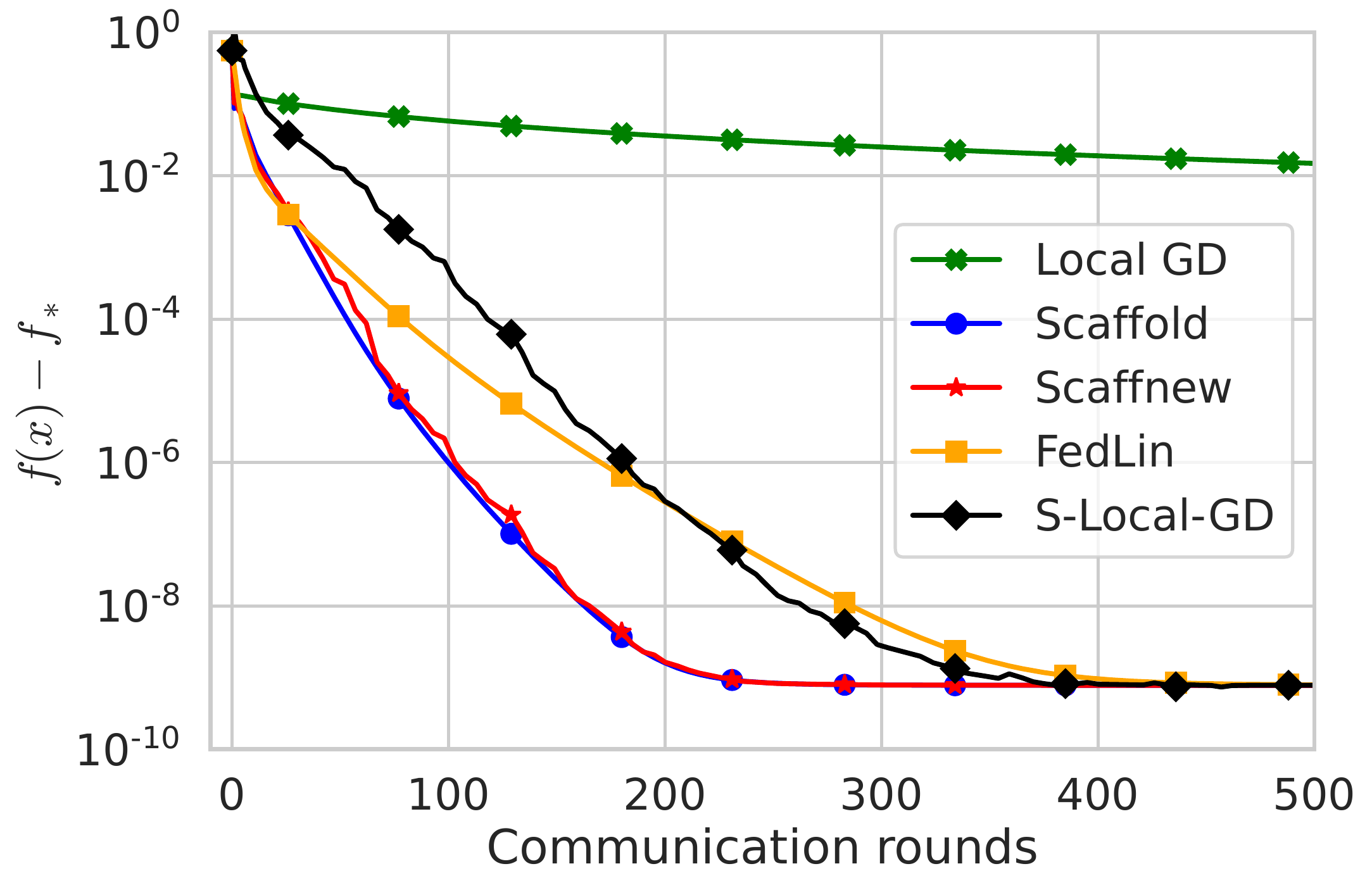}&

		$\!\!$\includegraphics[scale=0.25]{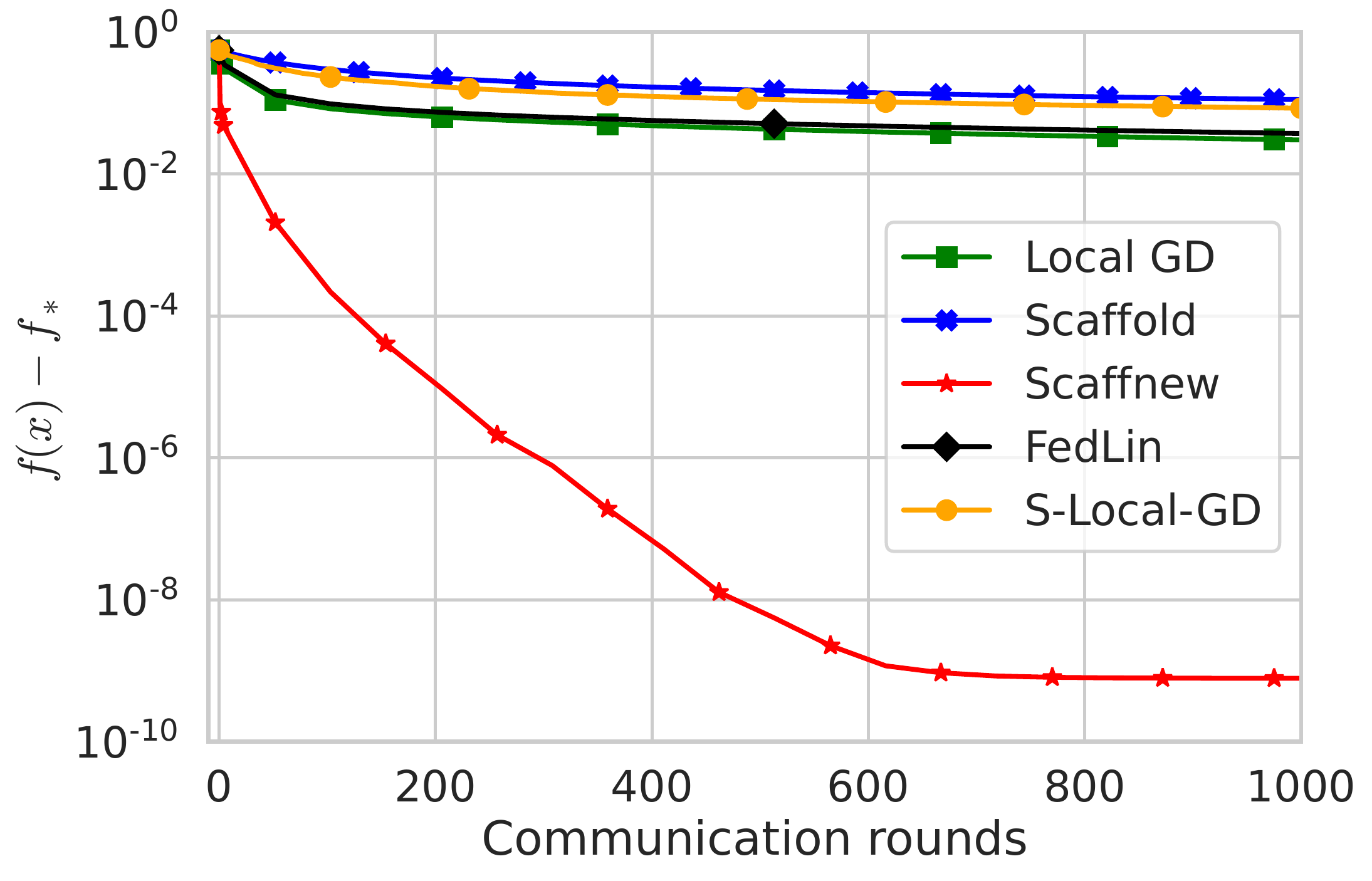}&
						$\!\!$\includegraphics[scale=0.25]{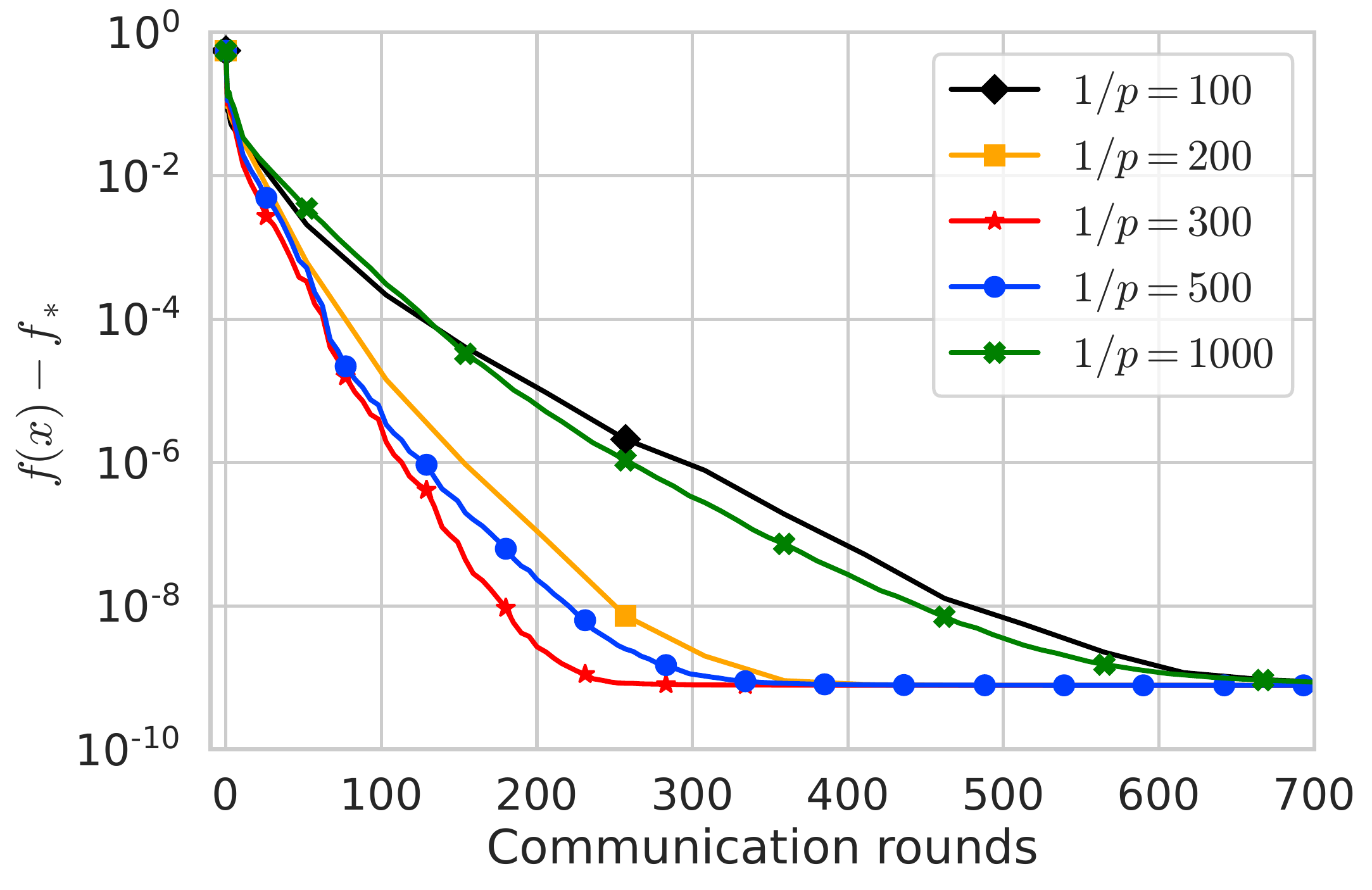}\\
		\small (a) tuned hyper-parameters & 
		\small (b) theoretical hyper-parameters&
		\small (c) different options of $p$
	\end{tabular}

	\caption{\textbf{Deterministic Case}. Comparison of \algname{Scaffnew} to other local update methods that tackle data-heterogeneity and to \algname{LocalGD}. In (a) we compare communication rounds with optimally tuned hyper-parameters. 
	In (b), we compare communication rounds with the algorithm parameters set to the best theoretical stepsizes used in the convergence proofs.
	In (c), we compare communication rounds with the algorithm stepsize set to the best theoretical stepsize and different options of parameter $p$.
	}
	\label{fig:image1}
\end{figure*}
\begin{figure*}[t]
	\centering
	\begin{tabular}{ccc}
		$\!\!$\includegraphics[scale=0.25]{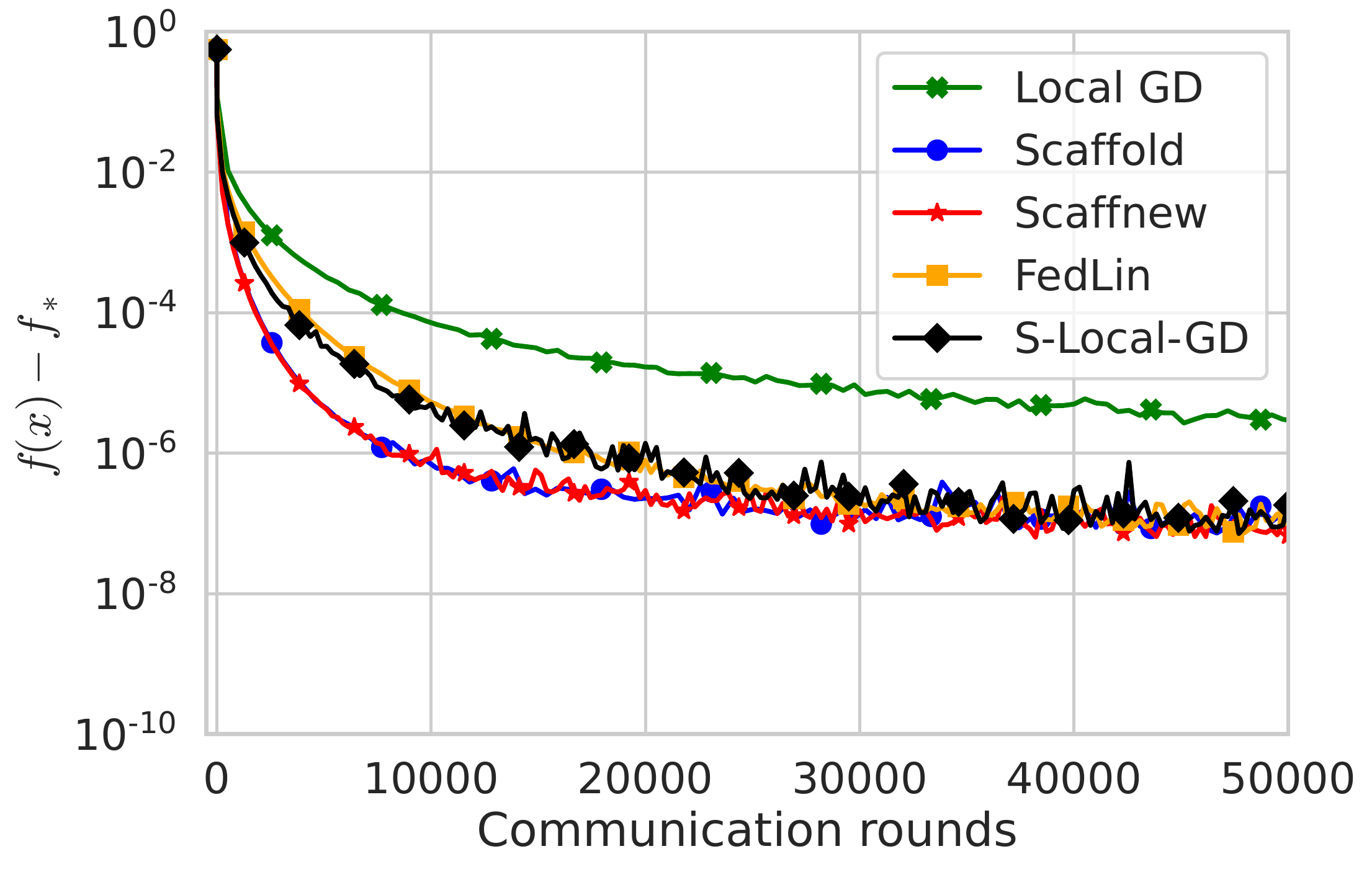}&
		$\!\!$\includegraphics[scale=0.25]{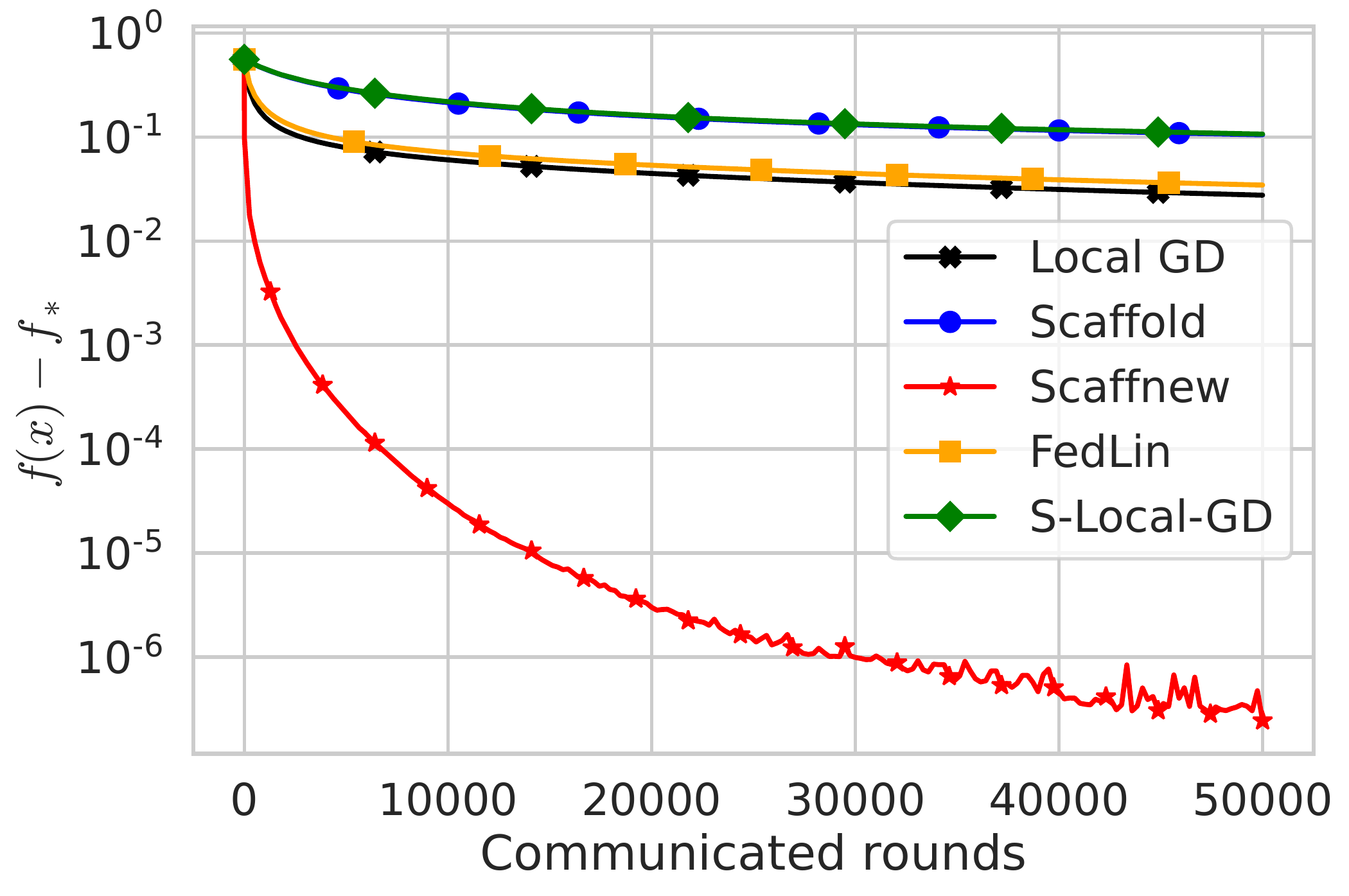}&
		$\!\!$\includegraphics[scale=0.25]{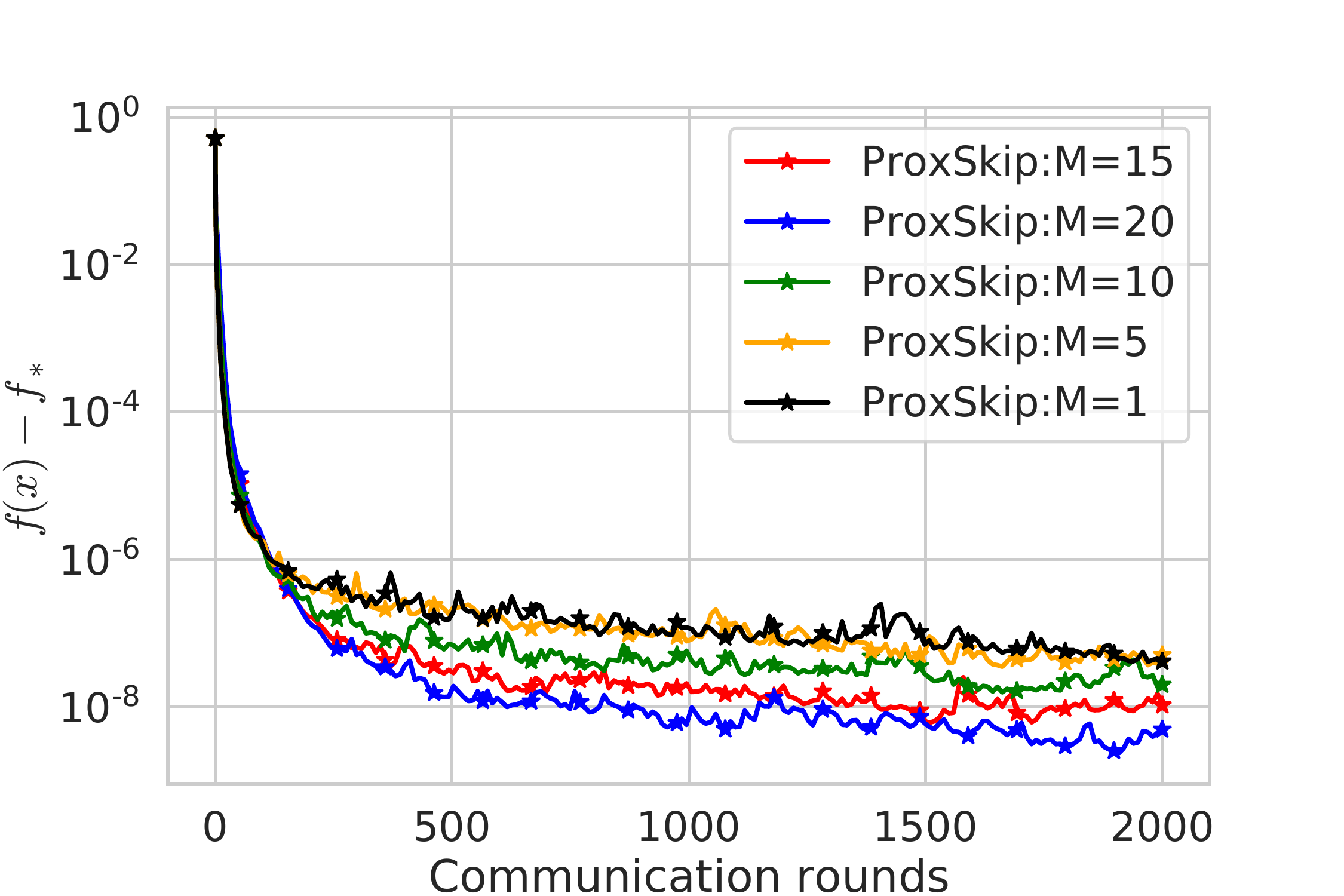}\\
		\small (a) tuned hyper-parameters & 
		\small (b) theoretical hyper-parameters&
			\small (c) different number of clients
	\end{tabular}
	
	\caption{\textbf{Stochastic Case}. Comparison of \algname{Scaffnew} to other local update methods that tackle data-heterogeneity  and to \algname{LocalSGD}. In (a) we compare commnication rounds with optimally tuned hyper-parameters. 
	In (b), we compare communication rounds with the algorithm parameters set to the best theoretical stepsizes used in the convergence proofs.
	In (c), we compare communication rounds with the algorithm parameters set to the best theoretical stepsizes used in the convergence proofs and different number of clients.
	}
	\label{fig:image2}
\end{figure*}

{\bf Limitations}. The main limitation of applying analysis of \algname{SProxSkip} in the FL setting (\Cref{alg:fl}) is that we do not achieve linear speedup in terms of the number of clients. This issue comes from the analysis technique and it needs deeper investigation. The same problem appears in the analysis of \algname{FedLin}, but it does not in the analysis of \algname{Scaffold}.

\subsection{Decentralized training}\label{sec:decentralized}

Let us now discuss the minimization problem with decentralized communication. Given a graph $G=(V, E)$ with $n$ nodes $V$ and edges $E$, we assume that every communication node $i$ receives a weighted average of its neighbors' vectors with weights $W_{i1},\dotsc, W_{in}\in [0, 1]$. Besides, nodes $i$ and $j$ communicate if and only if $W_{ij}\neq 0$, which is also equivalent to $(i, j)\in E$. The weights $W_{ij}$ define the \emph{mixing matrix} $\mW$ that we assume to be symmetric, doubly stochastic, and positive semi-definite. Then, the problem is equivalent to
\vspace{-1mm}
\[
\squeeze	
\min_{x\in\R^{d\cdot n}} f(x)\quad \text{subject to}\quad (\mI-\mW)x=0,
\]
where $\mI$ is the identity matrix. Let us set $\mL$ to be the square-root of $\mI-\mW$ and define the indicator function $\psi(y)$ by setting $\psi(0)=0$ and $\psi(y)=+\infty$ for any $y\neq 0$, which is similar to our previous definition in equation~\eqref{eq:constraint_consensus_0980}. Then, the constraint $(\mI-\mW)x=0$ is equivalent to $\mL x=0$, so the problem can be rewritten as
\begin{equation}
\squeeze	\min_{x\in\R^{d'}} f(x) + \psi(\mL x), \label{eq:problem_with_matrix}
\end{equation}
where $d'=d\cdot n$. The reason we define $\mL$ this way is that splitting algorithms require computation of $\mL\mL^\top u$ for some vector $u$, which in our case is exactly $(\mI-\mW)u$. Algorithmically, computing this product corresponds to communicating over the graph. For convenience, we provide the algorithm formulation in the graph notation in \Cref{alg:dist_gd}. The convergence of our decentralized algorithm is stated below.
\begin{theorem}\label{th:decentralized}
	Let $f$ satisfy Assumption \ref{as:f_i} and define the spectral gap of $\mW$ as $\delta=1-\lambda_2(\mW)\in(0, 1)$, where $\lambda_2(\mW)$ is the second largest eigenvalue of $\mW$. If we set $p\in(0, 1]$, $\gamma\le \nicefrac{1}{L}$, $\tau\le \nicefrac{p}{\gamma}$, then the average iterate $\overline x_T$ satisfies
	\[
		\E{\| \overline x_T-x_\star \|^2 } \le (1-\min(\gamma\mu, p\gamma\tau\delta))^T \Phi_0,
	\]
	where $\Phi_0\le \|x_0-x_\star\|^2 + \frac{\gamma}{p\tau\delta n}\sum_{i=1}^n\|\nabla f_i(x_*)\|^2$.
\end{theorem}
If we plug-in $\tau=\nicefrac{p}{\gamma}$, the theorem implies that the new rate is $\tilde{\cO}(\kappa + \frac{1}{p^2\delta})$. Thus, it is optimal to choose $p=\sqrt{\nicefrac{1}{(\delta\kappa)}}$ whenever the network is sufficiently well-connected. If passing a message is challenging, which happens when $\delta\le \nicefrac{1}{\kappa}$, then it is optimal to communicate every iteration by setting $p=1$. This trade-off is to be expected as our algorithm for \eqref{eq:problem_with_matrix} matches the lower bound of \citet{OPAPC} in terms of number of matrix-vector multiplications.
\section[Experiments]{Experiments}
To test the performance of algorithms and illustrate theoretical results, we use classical logistic regression problem. The loss function for this model has the following form:
\begin{align*}
	\squeeze 	f(x)=\frac{1}{N} \sum_{i=1}^{N} \log \left(1+\exp \left(-b_{i} a_{i}^{\top} x\right)\right)+\frac{\lambda}{2}\|x\|^{2},
\end{align*}
where $a_{i} \in \mathbb{R}^{d} \text { and } b_{i} \in\{-1,+1\}$ are the data samples and $N$ is their total number. We set the regularization parameter $\lambda = 10^{-4}L$, where $L$ is the smoothness constant.

 We implemented\footnote{Our code is available on GitHub: \url{https://github.com/alarcoelectro/ProxSkip-Public}} all algorithms in Python using the package RAY~\cite{moritz2018ray} to utilize parallelization. All methods were evaluated on a workstation with an Intel(R) Xeon(R) Gold 6146 CPU at 3.20GHz with 24 cores. We use the `w8a' dataset from LIBSVM library~\cite{chang2011libsvm}.
 
  In our experiments, we have two settings: deterministic (\Cref{fig:image1}) and stochastic problems (\Cref{fig:image2}). First, we provide results with tuned hyper-parameters (subplot (a)). \algname{Local GD} converges to the neighborhood of the solution due to data-heterogeneity. \algname{Scaffold} and \algname{Scaffnew} have the same convergence rate in terms of communication rounds and this rate is better than others. 
Second, we test algorithms with theoretical hyper-parameters (subplot (b)). In this setting, \algname{Scaffnew} outperforms other methods dramatically since our theory guarantees that we can use large stepsizes. The number of local steps is set to be $\sqrt{\hat{\kappa}}$, where $\hat{\kappa} = \frac{L}{\lambda}$.

 As we can see, if our method communicates either too often ($\nicefrac{1}{p} = 100$) or too rarely ($\nicefrac{1}{p} = 1000$), convergence suffers. The optimal number of local steps in this experiment is $\nicefrac{1}{p} = 300$. Our theory predicted that the choice $p=\frac{1}{\sqrt{\kappa}}$ is close to the experiment's results. Moreover, we compared Scaffnew in stochastic case with different number of clients $M$. As we can see, we can obtain the linear speedup, which is more optimistic than we have in theory. 

\section{Acknowledgements}
This work was supported by the French government under the management of the Agence Nationale de la Recherche as part of the "Investissements d'avenir" program, reference ANR-19-P3IA-0001 (PRAIRIE 3IA Institute).

{\small
\bibliography{scaffnew}
\bibliographystyle{icml2022-nourl} 
}

\newpage
\appendix
\onecolumn

\part*{Appendix}

\section{Basic Facts}
\label{appendix:facts}

The Bregman divergence of a differentiable function $f\colon\R^d \to \R$ is defined by
\[D_f(x,y) \eqdef f(x) - f(y) - \< \nabla f(y), x-y >. \]
It is easy to see that
\begin{equation} \label{eq:sym_Bregman} \< \nabla f(x) - \nabla f(y), x-y > = D_f(x,y) + D_f(y,x), \quad \forall x,y\in \R^d\end{equation}

For an $L$-smooth and  $\mu$-strongly convex function $f\colon\R^d\to \R$, we have
\begin{equation}\label{eq:bi87fgddf-1}\frac{\mu}{2} \|x-y\|^2 \leq D_f(x,y) \leq \frac{L}{2} \|x-y\|^2, \quad \forall x,y\in \R^d\end{equation}
and
\begin{equation}\label{eq:bi87fgddf-2}\frac{1}{2L} \|\nabla f(x)-\nabla f(y)\|^2 \leq D_f(x,y) \leq \frac{1}{2\mu} \|\nabla f(x)-\nabla f(y)\|^2, \quad \forall x,y\in \R^d.\end{equation}
Given $\psi\colon \R^d\to \R$, we define $\psi^*(y) \eqdef \sup_{x\in\R^d}\{\<x, y> - \psi(x)\}$ to be its Fenchel conjugate. The proximity operator of $\psi^*$ satisfies for any $\tau>0$
\begin{equation}
	\mathrm{if}\quad u=\prox_{\tau \psi^*}(y), \quad\mathrm{then}\quad u \in y - \tau\partial \psi^*(u). \label{eq:prox_implicit}
\end{equation}

\section{Analysis of \algname{ProxSkip}~(\Cref{alg:ProxSkip})}

\subsection{Proof of Lemma~\ref{lem:A}}

\begin{proof}
In order to simplify notation, let $P(\cdot)\eqdef\prox_{\frac{\gamma}{p}\psi}(\cdot)$, and  \begin{equation}\label{eq:x_and_y}x\eqdef\hat x_{t+1} - \frac{\gamma}{p}h_t, \qquad y\eqdef x_{\star} - \frac{\gamma}{p}h_{\star}.\end{equation}

{\bf STEP 1 (Optimality conditions).} Using the first-order optimality conditions for $f+\psi$ and using $h_{\star}\eqdef \nabla f(x_{\star})$, we obtain the following fixed-point identity for $x_{\star}$:
	\begin{equation}\label{eq:n08fhd90fd}
		x_{\star}
		= \prox_{\frac{\gamma}{p} \psi}\left(x_{\star} - \frac{\gamma}{p}h_{\star} \right) \overset{\eqref{eq:x_and_y}}{=} P(y).
	\end{equation}
	
{\bf STEP 2 (Recalling the steps of the method).}
Recall that the vectors $x_t$ and $h_t$ are in \Cref{alg:ProxSkip} updated as follows:
\begin{equation} \label{eq:step2a} x_{t+1} = \begin{cases} 
P\bigl(x \bigr) & \text{with probability} \quad p \\
 \hat x_{t+1} & \text{with probability} \quad 1- p 
\end{cases},\end{equation}
and
\begin{equation}\label{eq:step2b}  h_{t+1} = h_t + \frac{p}{\gamma}(x_{t+1} - \hat x_{t+1}) =\begin{cases}  h_t + \frac{p}{\gamma}(P(x)- \hat x_{t+1}) & \text{with probability} \quad p\\ h_t &  \text{with probability} \quad 1- p 
\end{cases}.
\end{equation}

{\bf STEP 3 (One-step expectation of the Lyapunov function).}

The expected value of the Lyapunov function 
\begin{equation}\label{eq:Lyapunov-proof} \Psi_t \eqdef \|x_{t} - x_{\star}\|^2 + \frac{\gamma^2}{p^2}\|h_{t} - h_{\star}\|^2\end{equation}
at time $t+1$, with respect to the coin toss at iteration $t$, is
{\footnotesize
\begin{eqnarray*}\E{\Psi_{t+1}} &\overset{\eqref{eq:step2a}+\eqref{eq:step2b}+\eqref{eq:Lyapunov-proof}}{=} & p \left(\|P(x) - x_{\star}\|^2 + \frac{\gamma^2}{p^2}\left\|h_t + \frac{p}{\gamma}(P(x)- \hat x_{t+1})- h_{\star} \right\|^2\right) + (1-p) \left(\| \hat x_{t+1}  - x_{\star}\|^2 + \frac{\gamma^2}{p^2}\|h_{t} - h_{\star}\|^2\right)\\
&\overset{\eqref{eq:n08fhd90fd}}{=}& p \left(\|P(x) - P(y)\|^2 + \left\| {\color{red}\frac{\gamma}{p}h_t + P(x)- \hat x_{t+1}} - {\color{blue} \frac{\gamma}{p}h_{\star} }\right\|^2\right) + (1-p) \left(\| \hat x_{t+1}  - x_{\star}\|^2 + \frac{\gamma^2}{p^2}\|h_{t} - h_{\star}\|^2\right)\\
&\overset{\eqref{eq:x_and_y}+\eqref{eq:n08fhd90fd}}{=} & p \left( \left\| P(x) - P(y) \right\|^2 + \underbrace{\left\| {\color{red}P(x) - x}   + {\color{blue} y-P(y)} \right\|^2}_{\|Q(x)-Q(y)\|^2} \right)  + (1-p) \left(\| \hat x_{t+1}  - x_{\star}\|^2 + \frac{\gamma^2}{p^2}\|h_{t} - h_{\star}\|^2\right).
\end{eqnarray*}
}

{\bf STEP 4 (Applying firm nonexpansiveness).}
Applying firm nonexpansiveness of $P$ (Lemma~\ref{lem:prox-contraction}), this leads to
 the inequality
\begin{eqnarray*}\E{\Psi_{t+1}} &\overset{\eqref{eq:prox_firm_non_exp}}{\le}  & p  \left\|x-y \right\|^2  + (1-p) \left(\| \hat x_{t+1}  - x_{\star}\|^2 + \frac{\gamma^2}{p^2}\|h_{t} - h_{\star}\|^2\right)\\
&\overset{\eqref{eq:x_and_y}}{=} & p \left\|\hat x_{t+1} - \frac{\gamma}{p}h_t - \left(x_{\star} - \frac{\gamma}{p}h_{\star} \right) \right\|^2 + (1-p) \left(\| \hat x_{t+1}  - x_{\star}\|^2 + \frac{\gamma^2}{p^2}\|h_{t} - h_{\star}\|^2\right).
\end{eqnarray*}

	
{\bf STEP 5 (Simple algebra).}	
Next, we expand the squared norm and collect the terms, obtaining
	\begin{eqnarray}
		\mathbb{E}\left[\Psi_{t+1} \right] &\leq & p\|\hat x_{t+1} - x_{\star}\|^2 + p\frac{\gamma^2}{p^2}\|h_t-h_{\star}\|^2 - 2\gamma\<\hat x_{t+1}-x_{\star}, h_t-h_{\star}> + (1-p)\Bigl( \|\hat x_{t+1} - x_{\star}\|^2 + \frac{\gamma^2}{p^2}\|h_{t} - h_{\star}\|^2\Bigr)  \notag \\
		&=& \|\hat x_{t+1} - x_{\star}\|^2 - 2\gamma\<\hat x_{t+1}-x_{\star}, h_t-h_{\star}> + \frac{\gamma^2}{p^2}\|h_t-h_{\star}\|^2. \label{eq:FINAL-A}
	\end{eqnarray}
Finally, note that by our definition of $w_t$, we have the identity $\hat x_{t+1} = w_t + \gamma h_t$. Therefore, the first two terms above can be rewritten as
	\begin{eqnarray}
		\|\hat x_{t+1} - x_{\star}\|^2 - 2\gamma\<\hat x_{t+1}-x_{\star}, h_t-h_{\star}>&=& \|w_{t} - w_{\star} + \gamma(h_t-h_{\star})\|^2 - 2\gamma\<w_{t}-w_{\star} + \gamma (h_t - h_{\star}), h_t-h_{\star}> \notag \\
		& =&  \|w_{t} - w_{\star}\|^2 + 2\gamma\<w_t-w_{\star}, h_t-h_{\star}> + \gamma^2 \|h_t-h_{\star}\|^2 \notag \\
		&&\qquad  - 2\gamma\<w_{t}-w_{\star}, h_t-h_{\star}> - 2\gamma^2 \|h_t - h_{\star}\|^2  \notag  \\
		& =& \|w_{t} - w_{\star}\|^2 - \gamma^2 \|h_t-h_{\star}\|^2. \label{eq:FINAL-B}
	\end{eqnarray}
	It remains to plug \eqref{eq:FINAL-B} into \eqref{eq:FINAL-A}.
\end{proof}

\subsection{Proof of Lemma~\ref{lem:B}}

\begin{proof}
	Recall the definition of $w_t$ and $w_{\star}$ in~\eqref{eq:98g9gbjfd8d}.
	Plugging these expressions into $\|w_t-w_{\star}\|^2$, expanding the square, and applying properties of $f$ as a $\mu$-strong convex and $L$-smooth function, we get 
	\begin{eqnarray*}
		\|w_{t} - w_{\star}\|^2&\overset{\eqref{eq:98g9gbjfd8d}}{=} &
		\|x_t - x_{\star} - \gamma (\nabla f(x_t)   - \nabla f(x_{\star}))\|^2 \\
		&=	&\|x_t - x_{\star}\|^2 + \gamma^2\|\nabla f(x_t) - \nabla f(x_{\star})\|^2 -2\gamma\<\nabla f(x_t) - \nabla f(x_{\star}), x_t - x_{\star}>\\
		&\overset{\eqref{eq:bi87fgddf-1}}{\le} & (1-\gamma\mu)\|x_t - x_{\star}\|^2 - 2\gamma  D_f(x_t,x_{\star}) +\gamma^2\|\nabla f(x_t) - \nabla f(x_{\star})\|^2 \\
		& = & (1-\gamma\mu)\|x_t - x_{\star}\|^2 - 2\gamma \left(D_f(x_t,x_{\star}) -  \frac{\gamma}{2} \|\nabla f(x_t) - \nabla f(x_{\star})\|^2 \right) \\
		& \overset{\eqref{eq:bi87fgddf-2}}{\le} &(1-\gamma\mu)\|x_t - x_{\star}\|^2,
	\end{eqnarray*}
	\normalsize
	where the last inequality holds if $0\leq \gamma \leq \frac{1}{L}$.
\end{proof}

\clearpage
\section{Analysis of \algname{SProxSkip}~(\Cref{alg:stoch_rand_prox})}

\subsection{The algorithm}

We consider a variant of \algname{ProxSkip} which uses a {\color{blue}{\em stochastic} gradient $g_t(x_t)$} instead of $\nabla f(x_t)$; see Algorithm~\ref{alg:stoch_rand_prox}.

\begin{algorithm*}[h]
	\caption{\algname{SProxSkip} (Stochastic gradient version of \algname{ProxSkip})}
	\label{alg:stoch_rand_prox}
	\begin{algorithmic}[1]
		\STATE stepsize $\gamma > 0$, probability $p>0$, initial iterate $x_0\in \R^d$, initial control variate ${\red h_0} \in \R^d$, number of iterations $T\geq 1$
		\FOR{$t=0,1,\dotsc,T-1$}
		\STATE $\hat x_{t+1} = x_t - \gamma ({\color{blue}g_t(x_t)} - {\red h_t})$ \hfill $\diamond$ Take a {\color{blue}stochastic} gradient-type step adjusted via the control variate ${\red h_t}$
		\STATE Flip a coin $\theta_t \in \{0,1\}$ where $\mathop{\rm Prob}(\theta_t =1) = p$ \hfill $\diamond$ Flip a coin that decides whether to skip the prox or not
		\IF{$\theta_t=1$} 
		\STATE  $x_{t+1} = \prox_{\frac{\gamma}{p}\psi}\bigl(\hat x_{t+1} - \frac{\gamma}{p}{\red h_t} \bigr)$ \hfill $\diamond$ Apply prox, but only very rarely! (with small probability $p$)
		\ELSE
		\STATE $x_{t+1} = \hat x_{t+1}$ \hfill $\diamond$ Skip the prox!
		\ENDIF
		\STATE ${\red h_{t+1}} = {\red h_t} + \frac{p}{\gamma}(x_{t+1} - \hat x_{t+1})$ \hfill $\diamond$ Update the control variate ${\red h_t}$
		\ENDFOR
	\end{algorithmic}
\end{algorithm*}

\subsection{Two lemmas}

Lemma~\ref{lem:A-stochastic} is an extension of Lemma~\ref{lem:A} to the stochastic case. In this result, we work with  \begin{equation}\label{eq:w'_t}w'_t = x_t - \gamma {\color{blue} g_t(x_t)} \end{equation} instead of $w_t = x_t - \gamma \nabla f(x_t)$.

\begin{lemma}\label{lem:A-stochastic}
If Assumptions \ref{as:f} and \ref{as:proper_psi} hold, $\gamma >0$ and $0<p\leq 1$, then
\begin{equation} \label{eq:b8f9d89fd8df_09}\squeeze \E{ \Psi_{t+1} } \leq  \|w'_t - w_{\star}\|^2 +  (1-p^2)\frac{\gamma^2}{p^2}\|h_{t} - h_{\star}\|^2 \,, \end{equation}
where the expectation is taken over the $\theta_t$ in \Cref{alg:stoch_rand_prox}.
\end{lemma}
\begin{proof} The proof is identical to the proof of Lemma~\ref{lem:A}. 
\end{proof}

Likewise, Lemma~\ref{lem:B-stochastic} is an extension of  Lemma~\ref{lem:B} to the stochastic case.

\begin{lemma} \label{lem:B-stochastic} Let \Cref{as:f} hold with any $\mu\geq 0$. If $0< \gamma \leq \frac{1}{A}$, then \begin{equation}\label{eq:nbo98fd8f_09uf00}\E{\|w'_{t} - w_{\star}\|^2} \le (1-\gamma\mu)\|x_t - x_{\star}\|^2 + \gamma^2C,\end{equation}
where the expectation is w.r.t.\ the randomness in the stochastic gradient ${\color{blue} g_t(\cdot)}$.
\end{lemma}

\begin{proof}
	Recall the definition of $w'_t$ in \eqref{eq:w'_t} and $w_{\star}$ in~\eqref{eq:98g9gbjfd8d}.
	Plugging these expressions into $\|w'_t-w_{\star}\|^2$, expanding the square, we get 
	\begin{eqnarray}
		\|w'_{t} - w_{\star}\|^2 &\overset{\eqref{eq:98g9gbjfd8d}+\eqref{eq:w'_t}}{=} &
		\|x_t - x_{\star} - \gamma (g_t(x_t)   - \nabla f(x_{\star}))\|^2 \notag \\
		&=	&\|x_t - x_{\star}\|^2 + \gamma^2\|{\color{blue} g_t(x_t)} - \nabla f(x_{\star})\|^2 -2\gamma\< {\color{blue} g_t(x_t)} - \nabla f(x_{\star}), x_t - x_{\star}>. \label{eq: bjhbidys_09}
	\end{eqnarray}
Taking expectation w.r.t.\ the randomness of the stochastic gradient $g_t(x_t)$, and using unbiasedness (\Cref{as:exp_smooth})  	and expected smoothness (\Cref{as:unbias}), we get	
	\begin{eqnarray*}
	\E{ \|w'_{t} - w_{\star}\|^2} & \overset{\eqref{eq: bjhbidys_09}}{=} & \|x_t - x_{\star}\|^2 + \gamma^2 \E{\|{\color{blue} g_t(x_t)} - \nabla f(x_{\star})\|^2} -2\gamma\< \E{{\color{blue} g_t(x_t)}} - \nabla f(x_{\star}), x_t - x_{\star}> \\		
	&=&  \|x_t - x_{\star}\|^2 -2\gamma\< \nabla f(x_t)- \nabla f(x_{\star}), x_t - x_{\star}> + \gamma^2 \E{\|{\color{blue} g_t(x_t)} - \nabla f(x_{\star})\|^2}  .	
	\end{eqnarray*}	

The second term can be decomposed using the identity  $\left\langle \nabla f(x_t) - \nabla f(x_{\star}) , x_t - x_{\star}\right\rangle = D_f(x_t,x_{\star})+D_f(x_{\star},x_t)$ (see \eqref{eq:sym_Bregman}), and the third term can be bounded via $\E{\|{\color{blue} g_t(x_t)} - \nabla f(x_{\star})\|^2} \leq 2AD_f(x_t,x_{\star})+C$ (see expected smoothness; \Cref{as:exp_smooth}), which leads to
\begin{equation}\label{eq:noihfd_9u0fd9}
	\E{\|w'_t - w_{\star}\|^2} \leq  \|x_t - x_{\star}\|^2 -  2\gamma ( D_f(x_t,x_{\star})+D_f(x_{\star},x_t))  + \gamma^2\left( 2AD_f(x_t,x_{\star})+C \right).
\end{equation}

By plugging the inequality $\mu\|x_t - x_{\star}\|^2\leq 2D_f(x_{\star},x_t)$ (see \eqref{eq:bi87fgddf-1}) into \eqref{eq:noihfd_9u0fd9}, we get
\begin{eqnarray*}
	\E{ \|w'_t - w_{\star}\|^2} &\leq & (1-\gamma\mu)\|x_t - x_{\star}\|^2  - 2\gamma  D_f(x_t,x_{\star}) + \gamma^2\left( 2AD_f(x_t,x_{\star})+C \right)\\
	&\leq & (1-\gamma\mu)\|x_t - x_{\star}\|^2  - 2\gamma (1-\gamma A)  D_f(x_t,x_{\star})+ \gamma^2C.
\end{eqnarray*}
Finally, the stepsize restriction $\gamma\leq \frac{1}{A}$ allows us to produce the estimate
\begin{equation}\label{eq:090fd9hf}
	\E{\|w'_t - w_{\star}\|^2}  \leq   (1-\gamma\mu) \|x_t - x_{\star}\|^2 + \gamma^2C,
\end{equation}
which is what we wanted to show.
\end{proof}

		

\subsection{Proof of Theorem~\ref{thm:main-stoch}}

\begin{proof}

Combining Lemma~\ref{lem:A-stochastic} and Lemma~\ref{lem:B-stochastic}, we get
\begin{eqnarray*}  \E{ \Psi_{t+1} } & \overset{\eqref{eq:b8f9d89fd8df_09}+\eqref{eq:090fd9hf}}{\leq} &  (1-\gamma\mu) \|x_t - x_{\star}\|^2  +  (1-p^2)\frac{\gamma^2}{p^2}\|h_{t} - h_{\star}\|^2 + \gamma^2C \\
&\leq & \max \{ 1-\gamma\mu, 1-p^2\} \Psi_t +\gamma^2 C\\
&=& (1-\zeta)\Psi_t +\gamma^2 C\, ,
 \end{eqnarray*}
 where $\zeta\eqdef \min\{\gamma\mu, p^2\}$. Taking full expectation, we get
$ \E{ \Psi_{t+1} } \leq  (1-\zeta) \E{\Psi_t} +\gamma^2 C $,
 and unrolling the recurrence, we finally obtain
 \begin{equation}\label{eq:yu8tgufd090f}\E{ \Psi_{T} } \leq (1 - \zeta)^T \Psi_0 + \frac{\gamma^2 C}{\zeta} \,.\end{equation}

\end{proof}

\subsection{Proof of Corollary~\ref{cor:0099887766}}
Recall that Theorem~\ref{thm:main-stoch} requires the stepsize $\gamma$ to satisfy \begin{equation} \label{eq:xx_one}0<\gamma \le \frac{1}{A}.\end{equation}
Pick $0<\varepsilon<1$. We will now choose $\gamma$ and $T$ such that $\E{ \Psi_{T} } \leq \varepsilon $. We shall do so by bounding both terms on the right-hand side of \eqref{eq:yu8tgufd090f}  by $\frac{\varepsilon}{2}$. 
\begin{itemize}
\item In order to minimize the number of prox evaluations, whatever the choice of $\gamma$ will be, we choose the smallest probability $p$ which does not lead to any degradation of the rate $\zeta\eqdef \min\{\gamma\mu, p^2\}$. That is, we choose \begin{equation}\label{eq:xx_p}p=\sqrt{\gamma \mu},\end{equation} in which case $\zeta = \gamma \mu$.
\item The first term on the right-hand side of \eqref{eq:yu8tgufd090f} can be bounded as follows:
\begin{equation}\label{eq:xx_three}T \geq \frac{1}{\gamma \mu} \log \left(\frac{2 \Psi_0}{\varepsilon}\right)  \quad \Longrightarrow \quad  (1 - \zeta)^T \Psi_0 \leq \frac{\varepsilon}{2}  .\end{equation}
\item The second term on the right-hand side of \eqref{eq:yu8tgufd090f} can be bounded as follows:
\begin{equation}\label{eq:xx_two}\gamma \leq \frac{ \varepsilon \mu  }{2 C} \quad \Longrightarrow \quad \frac{\gamma^2 C }{\zeta} = \frac{\gamma C }{\mu} \leq \frac{\varepsilon}{2} .\end{equation}

\end{itemize}

Since the number of iterations \eqref{eq:xx_three} depends inversely on the stepsize $\gamma$, we choose the largest stepsize consistent with the bounds \eqref{eq:xx_one} and \eqref{eq:xx_two}: \begin{equation}\label{eq:xx_four}\gamma = \min\left\{  \frac{1}{A}, \frac{ \varepsilon \mu}{2 C} \right\}.\end{equation}

By plugging this into \eqref{eq:xx_three}, we get the iteration complexity bound 
\[T \geq \max \left\{ \frac{A}{\mu}, \frac{2C}{\varepsilon \mu^2}\right\}\log \left(\frac{2 \Psi_0}{\varepsilon}\right) \quad \Longrightarrow \quad \E{ \Psi_{T} } \leq \varepsilon\,.\]

Since in each iteration we evaluate the prox with probability 
$p$ given by \eqref{eq:xx_p}, and since there are $T$ iterations, the expected number of prox evaluations is given by
\[p T \overset{\eqref{eq:xx_three}}{\geq} p \frac{1}{\gamma \mu} \log \left(\frac{2 \Psi_0}{\varepsilon}\right)  \overset{\eqref{eq:xx_p}}{=} \sqrt{\frac{1}{\gamma \mu}} \log \left(\frac{2 \Psi_0}{\varepsilon}\right) \overset{\eqref{eq:xx_four}}{=}\max\left\{ \sqrt{\frac{A}{\mu}}, \sqrt{\frac{2C}{\varepsilon \mu^2}}\right\} \log \left(\frac{2 \Psi_0}{\varepsilon}\right). \]

\clearpage

\section{Decentralized Analysis}

Let us now analyze the convergence of \Cref{alg:dist_gd} by introducing an algorithm for problem $\min_x f(x)+\psi(\mL x)$ and studying its properties.

\begin{algorithm*}[h]
	\caption{\algname{SplitSkip}}
	\label{alg:split_skip}
	\begin{algorithmic}[1]
		\STATE stepsizes $\gamma > 0$ and $\tau>0$, matrix $\mL\in\R^{m\times d}$,  probability $p>0$, initial iterate $x_0\in \R^d$, initial control variate ${\red y_0}=0 \in \R^m$, number of iterations $T\geq 1$
		\FOR{$t=0,1,\dotsc,T-1$}
		\STATE $\hat x_{t+1} = x_t - \gamma (\nabla f (x_t) +\mL^\top {\red y_t})$ \hfill $\diamond$ Take a gradient-type step adjusted via the control variate ${\red y_t}$
		\STATE Flip a coin $\theta_t \in \{0,1\}$ where $\mathop{\rm Prob}(\theta_t =1) = p$ \hfill $\diamond$ Flip a coin that decides whether to skip the prox or not
		\IF{$\theta_t=1$}
		\STATE  ${\red y_{t+1}} = \prox_{\tau\psi^*}\bigl( {\red y_{t}} +\tau \mL\hat{x}_{t+1} \bigr)$ \hfill $\diamond$ Apply prox, but only very rarely! (with small probability $p$)
		\STATE $x_{t+1} = \hat{x}_{t+1} - \frac{\gamma}{p}\mL^\top ({\red y_{t+1}} - {\red y_{t}})$
		\ELSE
		\STATE $x_{t+1} = \hat x_{t+1}$
		\STATE ${\red y_{t+1}} = {\red y_t}$ \hfill $\diamond$ Skip the prox!
		\ENDIF
		\ENDFOR
	\end{algorithmic}
\end{algorithm*}

 Notice that \Cref{alg:dist_gd} is a special case of \Cref{alg:split_skip} with $\mL=(\mI-\mW)^{1/2}$ and $\psi$ being indicator function of 0. Indeed, the conjugate of $\psi$ is simply 0 everywhere, so $\prox_{\tau\psi^*}(y)=y$ for any $y$. Thus, if we define $h_t \eqdef -\mL^\top y_t$, we obtain \Cref{alg:dist_gd}, as shown in more detail in Section~\ref{sec:formal_decentral}. For this reason, we will do the analysis for the more general \Cref{alg:split_skip}. 

 Let us also add a few connections of this method to the existing literature on primal--dual algorithms. When $p=1$, \Cref{alg:split_skip} reverts to a primal--dual algorithm first proposed by Loris and Verhoven for least squares problems~\cite{lor11}, and rediscovered later under the names \algname{PDFP2O}~\cite{chen2013primal} and \algname{PAPC}~\cite{DRORI2015209}. The convergence of this algorithm has been analyzed by~\citet{combettes2014forward,condat2019proximal,condat2022distributed} and generalized to the case of stochastic gradients by~\citet{salim2020dualize}, who also studied its linear convergence under similar assumptions, albeit without skipping the prox step.
 
 Our analysis is based on the following Lyapunov function:
\begin{align} 
 \Phi_t \eqdef \|x_{t} - x_{\star}\|^2 + \frac{\gamma}{p\tau}\|y_{t} - y_{\star}\|^2 , \label{eq:Lyapunov_decent}
\end{align}
where $y_t$ is the dual variable from \Cref{alg:split_skip}. 
\begin{theorem}\label{th:split_skip}
	Let \Cref{as:f} and \Cref{as:proper_psi} hold, and assume that for any $y$, we have $\partial\psi^*(y)\subseteq \range(\mL)$. If we choose $p\in(0, 1]$, $\gamma\le \frac{1}{L}$, $\gamma\tau \le \frac{p}{\|\mL\mL^\top\|}$, then
		\[
		\mathbb{E}\left[\|x_T-x_\star\|^2\right]
		\le (1 - \zeta)^T\Phi_0,
	\]
	where $\zeta=\min\{\gamma\mu, p\gamma\tau\lambda_{\min}^+(\mL\mL^\top)\}$.
\end{theorem}
\begin{proof}
	Let us define $\hat y_{t+1} \eqdef \prox_{\tau \psi^*}(y_t + \tau \mL\hat x_{t+1})$. As stated in equation~\eqref{eq:prox_implicit}, this definition implies the following implicit representation of $\hat y_{t+1}$:
	\[
		\hat y_{t+1}=y_t + \tau \mL\hat x_{t+1} - \tau (\psi^*) '(y_{t+1}),
	\]
	where $(\psi^*) '(y_{t+1})\in \partial \psi^*(y_{t+1})$ is a subgradient of $\psi^*$ at $y_{t+1}$.
	With the help of $\hat y_{t+1}$, we can expand the expected distance to the solution,
	\begin{align*}
		\mathbb{E}\left[\|x_{t+1} - x_\star\|^2\right]
		&= p \Bigl\|\hat x_{t+1} - x_\star - \frac{\gamma}{p}\mL^\top (\hat y_{t+1} - y_t)\Bigr\|^2 + (1-p)\|\hat x_{t+1} - x_\star\|^2 \\
		&= p \left[\|\hat x_{t+1} - x_\star\|^2 - 2\frac{\gamma}{p}\<\hat x_{t+1} - x_\star, \mL^\top (\hat y_{t+1} - y_t)> + \frac{\gamma^2}{p^2}\|\mL^\top (\hat y_{t+1} - y_t)\|^2\right] + (1-p)\|\hat x_{t+1} - x_\star\|^2 \\
		&= \|\hat x_{t+1} - x_\star\|^2 - 2\gamma \<\hat x_{t+1} - x_\star, \mL^\top (\hat y_{t+1} - y_t)> + \frac{\gamma^2}{p}\|\mL^\top (\hat y_{t+1} - y_t)\|^2.
	\end{align*}
	Next, let us recur the first term to $\|w_t- w_\star\|$ by using the expansion $\|a+b\|^2 = \|a\|^2 + 2\<a+b, b> - \|b\|^2$,
	\begin{align*}
		\|\hat x_{t+1} - x_\star\|^2
		&= \|w_t - w_\star - \gamma \mL^\top (y_t - y_\star)\|^2 \\
		&= \|w_t - w_\star\|^2 - 2\gamma \<\hat x_{t+1} - x_\star, \mL^\top (y_t - y_\star)> - \gamma^2\|\mL^\top (y_t - y_\star)\|^2.
	\end{align*}
	Thus,
	\begin{align*}
		\mathbb{E}\left[\|x_{t+1} - x_\star\|^2\right]
		&= \|w_t - w_\star\|^2 - 2\gamma \<\hat x_{t+1} - x_\star, \mL^\top (y_t - y_\star)> - 2\gamma \<\hat x_{t+1} - x_\star, \mL^\top (\hat y_{t+1} - y_t)> \\
		&\quad  - \gamma^2\|\mL^\top (y_t - y_\star)\|^2 + \frac{\gamma^2}{p}\|\mL^\top (\hat y_{t+1} - y_t)\|^2 \\
		&=\|w_t - w_\star\|^2 - 2\gamma \<\hat x_{t+1} - x_\star, \mL^\top (\hat y_{t+1} - y_\star)> - \gamma^2\|\mL^\top (y_t - y_\star)\|^2 + \frac{\gamma^2}{p}\|\mL^\top (\hat y_{t+1} - y_t)\|^2.
	\end{align*}
	Now, we can turn our attention to the convergence of the dual variable $y_t$. Since $y_{t+1}$ is updated with probability $p$, it holds
	\begin{align*}
		\mathbb{E}\left[\|y_{t+1} - y_{\star}\|^2\right]
		&= p \|y_t - y_\star + (\hat y_{t+1} - y_t)\|^2 + (1-p)\|y_t-y_\star\|^2 \\
		&= p \|y_t - y_\star\|^2 + 2p\<\hat y_{t+1}-y_\star,\hat y_{t+1} - y_t>  - p\|\hat y_{t+1} - y_t\|^2 + (1-p)\|y_t-y_\star\|^2 \\
		&= \|y_t - y_\star\|^2 + 2p\<\hat y_{t+1}-y_\star,\hat y_{t+1} - y_t>  - p\|\hat y_{t+1} - y_t\|^2 \\
		&= \|y_t - y_\star\|^2 + 2p\tau\<\hat y_{t+1}-y_\star,\mL \hat x_{t+1} - (\psi^*) '(\hat y_{t+1})>  - p\|\hat y_{t+1} - y_t\|^2.
	\end{align*}
	By the first-order optimality conditions, we have $\mL x_\star = (\psi^*)'(x_\star)$, where $(\psi^*)'(x_\star)$ is some subgradient of $\psi^*$ at $x_\star$. Therefore, convexity of $\psi^*$ gives
	\begin{align*}
		\<\hat y_{t+1}-y_\star, \mL \hat x_{t+1} - (\psi^*) '(\hat y_{t+1})>
		&= \<\hat y_{t+1}-y_\star,\mL (\hat x_{t+1} - x_\star) - (\psi^*) '(\hat y_{t+1}) + (\psi^*)'(y_\star)> \\
		&\le \<\hat y_{t+1}-y_\star,\mL (\hat x_{t+1} - x_\star)>.
	\end{align*}
	Combining the recursions for the iterates $x_{t+1}$ and $y_{t+1}$, we obtain
	\begin{align*}
		\mathbb{E}\left[\Phi_{t+1}\right]
		&= \mathbb{E}\left[\|x_{t+1}-x_\star\|^2 + \frac{\gamma}{p\tau}\|y_{t+1} - y_\star\|^2\right]\\
		&\le \|w_t - w_\star\|^2 - 2\gamma \<\hat x_{t+1} - x_\star, \mL^\top (\hat y_{t+1} - y_\star)> - \gamma^2\|\mL^\top (y_t - y_\star)\|^2 + \frac{\gamma^2}{p}\|\mL^\top (\hat y_{t+1} - y_t)\|^2\\
		&\quad + \frac{\gamma}{p\tau}\|y_t - y_\star\|^2 + 2\gamma\<\hat y_{t+1}-y_\star,\mL (\hat x_{t+1} - x_\star)>  - \frac{\gamma}{\tau}\|\hat y_{t+1} - y_t\|^2 \\
		&= \|w_t - w_\star\|^2 +  \frac{\gamma}{p\tau}\|y_t - y_\star\|^2  - \gamma^2\|\mL^\top (y_t - y_\star)\|^2 + \frac{\gamma^2}{p}\|\mL^\top (\hat y_{t+1} - y_t)\|^2  - \frac{\gamma}{\tau}\|\hat y_{t+1} - y_t\|^2.
	\end{align*}
	Using the assumption that $\tau \le \frac{p}{\gamma\|\mL \mL^\top\|}$, we get
	\[
		\frac{\gamma^2}{p}\|\mL^\top (\hat y_{t+1} - y_t)\|^2
		\le \frac{\gamma^2}{p}\|\mL\mL^\top\| \|\hat y_{t+1} - y_t\|^2
		\le \frac{\gamma}{\tau}\|\hat y_{t+1} - y_t\|^2.
	\]
	Plugging this back, we derive
	\[
		\mathbb{E}\left[\Phi_{t+1}\right]
		\le \|w_t - w_\star\|^2 +  \frac{\gamma}{p\tau}\|y_t - y_\star\|^2 - \gamma^2\|\mL^\top (y_t - y_\star)\|^2
		\overset{\eqref{eq:nbo98fd8f_09uf}}{\le} (1-\gamma\mu)\|x_t - x_*\|^2 +  \frac{\gamma}{p\tau}\|y_t - y_\star\|^2 - \gamma^2\|\mL^\top (y_t - y_\star)\|^2.
	\]
	Since we assume that $\partial \psi^*(y)\subseteq \range(\mL)$ and $y_0=0\in\R^d$, we have that $y_{t}-y_\star\in \range(\mL)$. Therefore, $\|\mL^\top (y_{t}-y_\star)\|^2\ge \lambda_{\min}^+(\mL\mL^\top)\|y_{t}- y_\star\|^2$, where $\lambda_{\min}^+$ is the smallest positive eigenvalue. Combining these results, we get
	\begin{align*}
		\mathbb{E}\left[\|x_t-x_\star\|^2\right]
		&\le \mathbb{E}\left[\Phi_t\right]
		\le (1-\gamma\mu)\|x_{t-1} - x_\star\|^2 +  \frac{\gamma}{p\tau}\left(1 - p\gamma\tau\lambda_{\min}^+(\mL\mL^\top)\right)\|y_{t-1} - y_\star\|^2 \\
		&\le (1 - \min(\gamma\mu, p\gamma\tau\lambda_{\min}^+(\mL\mL^\top))^t\Phi_0.  \qedhere
	\end{align*}
\end{proof}
\subsection{Proof of \Cref{th:decentralized}}\label{sec:formal_decentral}
\begin{proof}
	To obtain the communication step as a special case of the proximity operator, we set $\psi$ to be the indicator function of the singleton $\{0\}\subseteq \R^d$,
	\[
		\psi(x)= \begin{cases} 0 & x = 0 \\ +\infty & x\neq 0\end{cases}.
	\]
	Its Fenchel conjugate equals, by definition, $\psi^*(y)=\sup_{x\in\R^d}\{\<x, y> - \psi(x)\} = \<0, y>=0$. Therefore, for any $y$, $\prox_{\tau \psi^*}(y)=y$ and $\partial \psi^*(y)=\{0\}\subseteq \range(\mL)$, and the conditions of \Cref{th:split_skip} hold. Next, let us establish that \Cref{alg:dist_gd} is a special case of \Cref{alg:split_skip}. If we consider the iterates of \Cref{alg:split_skip} and define $h_t \eqdef -\mL^\top y_t$, then its first step can be rewritten as
	\[
		\hat x_{t+1} 
		= x_t - \gamma (\nabla f(x_t) + \mL^\top y_t)
		= x_t - \gamma (\nabla f(x_t) - h_t),
	\]
	which is exactly the first step of \Cref{alg:dist_gd}. The second step of \Cref{alg:split_skip} is either to do nothing or to update $y_{t+1}$. Using the fact that $\prox_{\tau \psi^*}(y_t + \tau \mL \hat x_{t+1})= y_t + \tau \mL \hat x_{t+1}$, it is easy to see that
	\[
		h_{t+1}
		\eqdef -\mL^\top y_{t+1}
		= -\mL^\top(y_t + \tau \mL \hat x_{t+1})
		= h_t - \tau \mL^\top \mL \hat x_{t+1}.
	\]
	By setting $\mL=(\mI-\mW)^{1/2}$, we get $\mL^\top \mL = \mI - \mW$, and we recover the second step of \Cref{alg:dist_gd} in an equivalent form:
	\[
	\begin{cases}
		h_{i,t+1} &= h_{i,t} + \tau\bigl( \hat x_{i, t+1} - \sum_{j=1}^n W_{ij}\hat x_{j,t+1}\bigr), \\
		x_{i, t+1} &= \hat x_{i, t+1} + \frac{\gamma}{p}(h_{i, t+1} - h_{i, t}).
	\end{cases}
	\qquad\Longleftrightarrow\qquad
	\begin{cases}
		x_{i,t+1} &= \left(1 - \frac{\gamma\tau}{p}\right)\hat x_{i,t+1} + \frac{\gamma\tau}{p}\sum_{j=1}^n W_{ij}\hat x_{j,t+1}, \\
		h_{i,t+1} &= h_{i,t} + \frac{p}{\gamma}(x_{i,t+1} -  \hat x_{i, t+1}).
	\end{cases}
	\]
	Finally, notice that $\lambda_{\min}^+(\mL^\top \mL) = \lambda_{\min}^+(\mI - \mW) = 1-\lambda_2(\mW) = \delta$, $\|\mL^\top \mL\| = \|\mI - \mW\| < 1$, and $y_{i,0}=0$, so applying \Cref{th:split_skip} yields
	\[
		\mathbb{E}\left[\frac{1}{n}\sum_{i=1}^n \|x_{i,T}-x_\star\|^2 \right]
		\le (1 - \zeta)^T\Phi_0
		= (1 - \zeta)^T\left(\|x_0 - x_\star\|^2 + \frac{\gamma}{p\tau}\frac{1}{n}\sum_{i=1}^n\|y_{i,*}\|^2 \right).
	\]
	By Jensen's inequality, the average iterate $\overline x_T \eqdef \frac{1}{n}\sum_{i=1}^n x_{i, T}$ satisfies
	\[
		\mathbb{E}\left[\|\overline x_T - x_\star\|^2 \right]
		\le \mathbb{E}\left[\frac{1}{n}\sum_{i=1}^n \|x_{i,T}-x_\star\|^2 \right]
		\le (1 - \zeta)^T \left(\|x_0 - x_\star\|^2 + \frac{\gamma}{p\tau}\frac{1}{n}\sum_{i=1}^n\|y_{i,*}\|^2 \right).
	\]
	Finally, notice that $\|y_{i,*}\|^2 = \|\mL^{\dagger}\mL y_{i,*}\|^2 = \|\mL^{\dagger}\nabla f_i(x_*)\|^2\le \frac{1}{\lambda_{\min}^+(\mL^\top \mL)}\|\nabla f_i(x_*)\|^2= \frac{1}{\delta}\|\nabla f_i(x_*)\|^2$.
\end{proof}

\clearpage

\end{document}